\documentclass[journal]{IEEEtran}

\usepackage{graphicx}
\usepackage{subfig}
\usepackage{amsmath,array}
\usepackage{amsfonts}
\usepackage{algorithm,algorithmic}
\usepackage{booktabs,multirow}
\usepackage{enumerate}
\usepackage{url}
\usepackage{hyperref}
\usepackage{color,soul}

\begin{document}

\title{\huge{Revealing the Invisible with Model and Data Shrinking for Composite-database Micro-expression Recognition}}

\author{Zhaoqiang~Xia, Wei~Peng, Huai-Qian~Khor, Xiaoyi~Feng, Guoying~Zhao
\thanks{Z. Xia is with School of Electronics and Information, Northwestern Polytechnical University, and also affiliated with Center for Machine Vision and Signal Analysis, University of Oulu. e-mail:xiazhaoqiang@gmail.com.}
\thanks{ X. Feng is with School of Electronics and Information, Northwestern Polytechnical University, 710129 Shaanxi.}
\thanks{H.Q. Khor, W. Peng and G. Zhao is with Center for Machine Vision and Signal Analysis, University of Oulu, 90014 Oulu, Finland.}
}


\maketitle

\begin{abstract}
Composite-database micro-expression recognition is attracting increasing attention as it is more practical to real-world applications. Though the composite database provides more sample diversity for learning good
representation models, the important subtle dynamics are prone to disappearing in the domain shift such that the models greatly degrade their performance, especially for deep models. In this paper, we analyze the influence of learning complexity, including the input complexity and model complexity, and discover that the lower-resolution input data and shallower-architecture model are helpful to ease the degradation of deep models in composite-database task. Based on this, we propose a recurrent convolutional network (RCN) to explore the shallower-architecture and lower-resolution input data, shrinking model and input complexities simultaneously. Furthermore, we develop three parameter-free modules (i.e., wide expansion, shortcut connection and attention unit) to integrate with RCN without increasing any learnable parameters. These three modules can enhance the representation ability in various perspectives while preserving not-very-deep architecture for lower-resolution data. Besides, three modules can further be combined by an automatic strategy (a neural architecture search strategy) and the searched architecture becomes more robust. Extensive experiments on MEGC2019 dataset (composited of existing SMIC, CASME II and SAMM datasets) have verified the influence of learning complexity and shown that RCNs with three modules and the searched combination outperform the state-of-the-art approaches.
\end{abstract}


\IEEEpeerreviewmaketitle

\section{Introduction}
\IEEEPARstart{M}{icro-expressions} occurred on human's faces provide an important visual clue for perceiving their genuine emotions while macro-expressions (normal expressions) sometimes can be pretended to hide the human's emotions. Thus, the micro-expressions have many potential applications in the fields of capturing the real psychological activities \cite{Takalkar2018A}, e.g., psychoanalysis, lie detection, criminal interrogation, and business negotiation. Due to the short-duration and subtle intensity changes of facial regions in the task of recognizing micro-expressions, it becomes very challenging for both humans and machines. Fueled by the success of computer vision techniques for macro-expression, the researchers attempt to recognize micro-expressions automatically and the task of micro-expression recognition (MER) becomes attractive after several benchmark datasets (e.g., SMIC \cite{li2013spontaneous}, CASME II \cite{yan2014casme} and SAMM \cite{Davison2018SAMM}) are publicly available.

For addressing the problem of MER, many approaches including conventional and deep methods have been developed to model the fleeting subtle changes of spontaneous micro-expressions towards the individual-database task. The conventional methods \cite{pfister2011recognising, Ruizhernandez2013Encoding, Wang2014LBP, Davison2014Micro, Huang2015Facial, Wang2015Rec, Wang2015Micro, Huang2016Spontaneous, Duan2016Recognizing, Liu2016A, Li2017Towards, Huang2017Discriminative, Liong2018Less, Happy2019Fuzzy} usually extract handcrafted features, e.g., local binary patterns on three orthogonal planes (LBP-TOP) \cite{pfister2011recognising,Davison2014Micro}, second-order Gaussian jet on LBP-TOP \cite{Ruizhernandez2013Encoding}, LBP six intersection points (LBP-SIP) \cite{Wang2014LBP}, local spatiotemporal directional features (LSDF) \cite{Wang2015Rec}, spatiotemporal LBP (STLBP) \cite{Huang2015Facial}, spatiotemporal completed local quantization patterns (STCLQP) \cite{Huang2016Spontaneous}, discriminative spatiotemporal LBP (DSLBP) \cite{Huang2017Discriminative}, directional mean optical-flow (MDMO) \cite{Liu2016A}, bi-weighted oriented optical flow (Bi-WOOF) \cite{Liong2018Less} and fuzzy histogram of optical flow orientation (FHOFO) \cite{Happy2019Fuzzy}, and then construct a classifier, e.g., support vector machine (SVM) \cite{pfister2011recognising, Wang2015Micro, Huang2015Facial, Huang2016Spontaneous, Liong2018Less} and random forest (RF) \cite{pfister2011recognising, Davison2014Micro, Duan2016Recognizing, Li2017Towards}, specially for MER. Although these handcrafted features continue to improve the representation ability for MER, it is still difficult to manually design good representations for capturing quick subtle changes of micro-expressions. Oppositely, deep convolutional neural networks (CNNs) can obtain high-quality descriptions automatically and outperform the conventional methods \cite{Zhang2019Deep}. Therefore, some deep models  such as the CNNs and long short-term memory (LSTM) \cite{Wang2018Micro}, pretrained CNNs (e.g., MagGA/SA \cite{Li2018Can} and OFF-ApexNet \cite{Gan2019Off}) and STRCN \cite{Xia2018Spontaneous, Xia2019Spatiotemporal} have been devoted to MER for improving the representation ability. These deep approaches model the spatio-temporal changes of micro-expressions and learn the visual features as well as the classifier in an end-to-end way. With the successful application of deep models, the recognition performance on individual datasets has been promoted greatly and reaches state-of-the-art results. However, these approaches towards the individual datasets cannot be applied successfully to real-world environments due to their strict conditions.

In real-world applications, it is possible to obtain micro-expression samples in multiple scenarios and labeled by various professional experts, which can collect more samples and have high diversity. This can be simulated by a composite-database task of MER and becomes more important than individual-database task due to its practicality. However, the composite-database task is more challenging as it would induce the domain shift between different scenarios (datasets). This would be aggravated when the samples of micro-expressions are limited. The performance of deep models is greatly degraded as the domain shift among mixed datasets disturbs the model learning without sufficient training samples, which has been exhibited in recent attempts \cite{Liong2019Shallow, Liu2019Neural, Zhou2019Dual, Van2019Capsulenet, Xia2019Cross} for micro-expression grand challenging (MEGC2019\footnote{https://facial-micro-expressiongc.github.io/MEGC2019/}). In this challenge, various architectures and different strategies for reducing domain noises were presented. However, even the deeper models are with more powerful representation \cite{Van2019Capsulenet, Xia2019Cross}, they cannot well deal with domain shift and harass the model focusing on the real description for MER. Motivated by this contradiction, in this paper, we analyze the influence of learning complexity mainly including the input complexity and model complexity. The input complexity usually includes the data number and data resolution while data number is always the focus of MER. So, in the context, the model complexity refers to the architecture layers while the input complexity means the input data resolution of feeding into the deep model. With qualitative and quantitative investigation in Section \ref{sec:pavm}, we find that the lower-resolution input data and shallower-architecture model are helpful to ease the degradation of deep models in composite-database task.

Based on this observation, we propose a recurrent convolutional network (RCN) and extend it with three parameter-free modules and a searchable combination to promote the representation ability by shrinking model and input complexities simultaneously. The proposed RCN is a shallower architecture and is fed with lower-resolution data, which provides lower learning complexity. To further promote the representation ability while avoiding the overfitting to domain shift, we develop three parameter-free extension modules, i.e., wide expansion, shortcut connection and attention unit, and integrate them into our RCN. These three modules will not increase any learnable parameters to RCN when integrating into the deep framework. So, the three modules can enhance the representation ability in various perspectives while preserving not-very-deep architecture for lower-resolution data. The low-complexity model and input will greatly improve the learning efficiency and save the computational cost, including the storage cost and time cost. Besides, three modules can be further combined by a neural architecture search (NAS) strategy, which efficiently explores a better combination of these modules and thus it can achieve more robust representation. Our main contributions in this context can be summarized as follows:
\begin{itemize}
  \item We analyze the influence of lower-resolution input data along with shallower-architecture models and reveal that they are helpful for easing the degradation of deep models in the composite-database MER task.
  \item A recurrent convolutional network is proposed with shallower architecture and lower-resolution input for restricting domain shift and improving learning efficiency.
  \item We develop three parameter-free modules and introduce a NAS search strategy to search an
optimal combination way for obtaining more robust representation without increasing learning complexity.
  \item Extensive experiments show that the proposed methods can achieve the superior performance compared to the state-of-the-art MER approaches.
\end{itemize}

The rest of this paper is organized as follows. The related work is briefly summarized in Section \ref{sec:rw} and the influence of learning complexity is analyzed in Section \ref{sec:pavm}. Our proposed approach for composite-database MER is presented in Section \ref{sec:md}. Then we discuss the experimental results for algorithm evaluation in Section \ref{sec:ex}. Finally, Section \ref{sec:con} concludes this work.

\section{Related Work}
\label{sec:rw}
In this section, we briefly review the individual-database, cross-database and composite-database micro-expression recognition (MER) approaches, respectively. The spontaneous micro-expression datasets usually used in MER and related micro-expression spotting approaches can be accessed in recent literature \cite{Takalkar2018A,Xia2016Spontaneous}.

\subsection{Individual-database MER}
The approaches of spontaneous MER on individual datasets are roughly divided into two types: \emph{conventional methods} and \emph{deep methods}. Conventional methods utilize handcrafted visual features and combine them with conventional classifiers while deep methods learn the visual feature and classifier in an end-to-end way.

\subsubsection{Conventional Methods}
In conventional methods, the appearance based features have been widely used. LBP-TOP features \cite{pfister2011recognising} calculating LBP codes from three orthogonal planes were combined with SVMs, multiple kernel classifiers or random forests respectively for MER. In the other work, LBP-TOP was extended into the tensor independent color space for obtaining more robust subspace \cite{wang2014micro,Wang2015Micro}. Furthermore, based on LBP-TOP, some extended methods \cite{Wang2014LBP, Wang2015Rec, Huang2015Facial, Huang2016Spontaneous, Huang2017Discriminative,Peng2019Boost} were proposed to improve the robustness and representation ability of LBP-TOP. LBP-SIP provided a lightweight representation based on three intersecting lines crossing over the center point of LBP-TOP \cite{Wang2014LBP}. LSDF \cite{Wang2015Rec} further used regions of interest to extract local directional features, which encoded the sign feature with magnitude information as weights. STCLBP \cite{Huang2015Facial}, STCLQP \cite{Huang2016Spontaneous} and DSLBP \cite{Huang2017Discriminative} extended LBP-TOP by containing more information (i.e., magnitude, orientation and shape attributes). Moreover, a hierarchical encoding way for LBP-TOP based features \cite{Zong2018Learning} was presented to consider multiple blocks of LBP-TOP.

On the contrary, geometric based features were used to extract motion deformations by using landmarks or optical flows. Facial landmarks \cite{yao2014micro} were firstly modeled to recognize limited-type micro-expressions (i.e., happiness and disgust). The Delaunay triangulation of several landmarks was used to capture subtle muscle movements for MER \cite{Lu2015Delaunay}. On the other side, based on optical flow estimation, some approaches leverage the magnitude, orientation and other high-order statistics to model the dynamics of micro-expressions. The MDMO \cite{Liu2016A} calculated histograms of ROIs by counting the oriented optical flow vectors while facial dynamics map (FDM) \cite{Xu2017Microexpression} calculated the direction histogram of spatiotemporal cuboids based on the micro-expression sequence. Similar to LSDF, Bi-WOOF \cite{Liong2018Less} used a flow orientation histogram to replace the LBP histogram by considering the magnitude and optical strain values as the weights while FHOFO \cite{Happy2019Fuzzy} fuzzily encoded the temporal pattern of optical flow orientation by considering the circular continuity of angular histograms during fuzzification.

\subsubsection{Deep Methods}
In recent two years, several deep methods have been proposed. The pretrained CNN \cite{Takalkar2017Image} was transferred and fine-tuned to recognize image based micro-expressions, in which each image (frame) in a video sequence was recognized for micro-expressions. In \cite{Wang2018Micro}, a pretrained CNN was first fine-tuned to extract visual features for each frame. Following CNN, long short-term memory (LSTM) was used to model the correlation of frames in one sequence and then capture the texture difference of entire sequence. Besides, only based on one apex frame detected from a sequence, VGG-Face was fine-tuned to recognize micro-expressions in \cite{Li2018Can}. Similarly, OFF-ApexNet \cite{Gan2019Off} extracted the optical flows of two directions respectively based on the onset frame and apex frame in one sequence and combined these two streams in a fully-connected layer for MER. In \cite{Khor2019Dual}, a deep model containing two-stream and two-level convolutional layers based on pretrained AlexNet was used to promote OFF-ApexNet and then fine tuned for recognizing micro-expressions. Different from using the pretrained models, STRCN \cite{Xia2018Spontaneous,Xia2019Spatiotemporal} modeled the spatio-temporal changes both in appearance based and geometric based ways, which can train the deep models from scratch by using data augmentation techniques. Due to the powerful representation ability of deep models, they have become the mainstream approaches in MER task.

\subsection{Cross-database MER}
Recently, there have been more attempts to introduce new works for cross-database MER. The cross-database task chooses the training samples from one dataset while the test samples should be chosen in another dataset, which is similar to a 2-fold cross validation in dataset level of individual datasets. Some works have been devoted to this type of cross-database problem. For instance, the subspace techniques used in face and expression recognition have been applied to the MER task, such as target sample re-generator (TSRG) \cite{Zong2018Domain} and transductive transfer regression model (TTRM) \cite{Zong2019Toward}. In the first micro-expressions grand challenge (MEGC2018), five approaches \cite{Yap2018Facial} using conventional methods and deep methods were compared on the five-class cross-database MER of CASME II and SAMM datasets. Among them, the deep method used a pretrained ResNet10 model achieved the best performance and outperformed other conventional methods according to the challenge report \cite{Yap2018Facial}.

\subsection{Composite-database MER}
Different from individual and cross-database tasks, the composite-database task mixes all samples of all datasets together and tests by choosing samples from the composite dataset. Except the cross-database task, MEGC2018 also reported the composite-database recognition results on the mixture of two datasets (CASME II and SAMM) \cite{Yap2018Facial}. Further, MEGC2019 extended it to three datasets (CASME II, SMIC and SAMM) and  only focused on composite-database task as it is more common in realistic environments compared to cross-database task. Several composite-database approaches have been presented \cite{Liong2019Shallow, Liu2019Neural, Zhou2019Dual, Van2019Capsulenet, Xia2019Cross} for MEGC2019 task. In these approaches, various deep models and strategies (e.g., two-stream or three-steam networks \cite{Zhou2019Dual, Liong2019Shallow}, capsule network \cite{Van2019Capsulenet}, transfer domain knowledge \cite{Liu2019Neural}, and sample normalization \cite{Xia2019Cross}) were presented to improve the robustness to the domain noises. Most approaches extracted the optical flow from one frame (apex or middle frame) to learn representations for three-class recognition while only \cite{Van2019Capsulenet} chose the apex frame (image) directly as model input for recognition. Different from previous works, in this paper, we propose recurrent convolutional networks with extension modules, which require no extra learning,  to further pursue more powerful representation ability as well as avoiding the disturbance of domain shift.

\begin{figure}[t]
  \centering
  \includegraphics[width=0.65\linewidth]{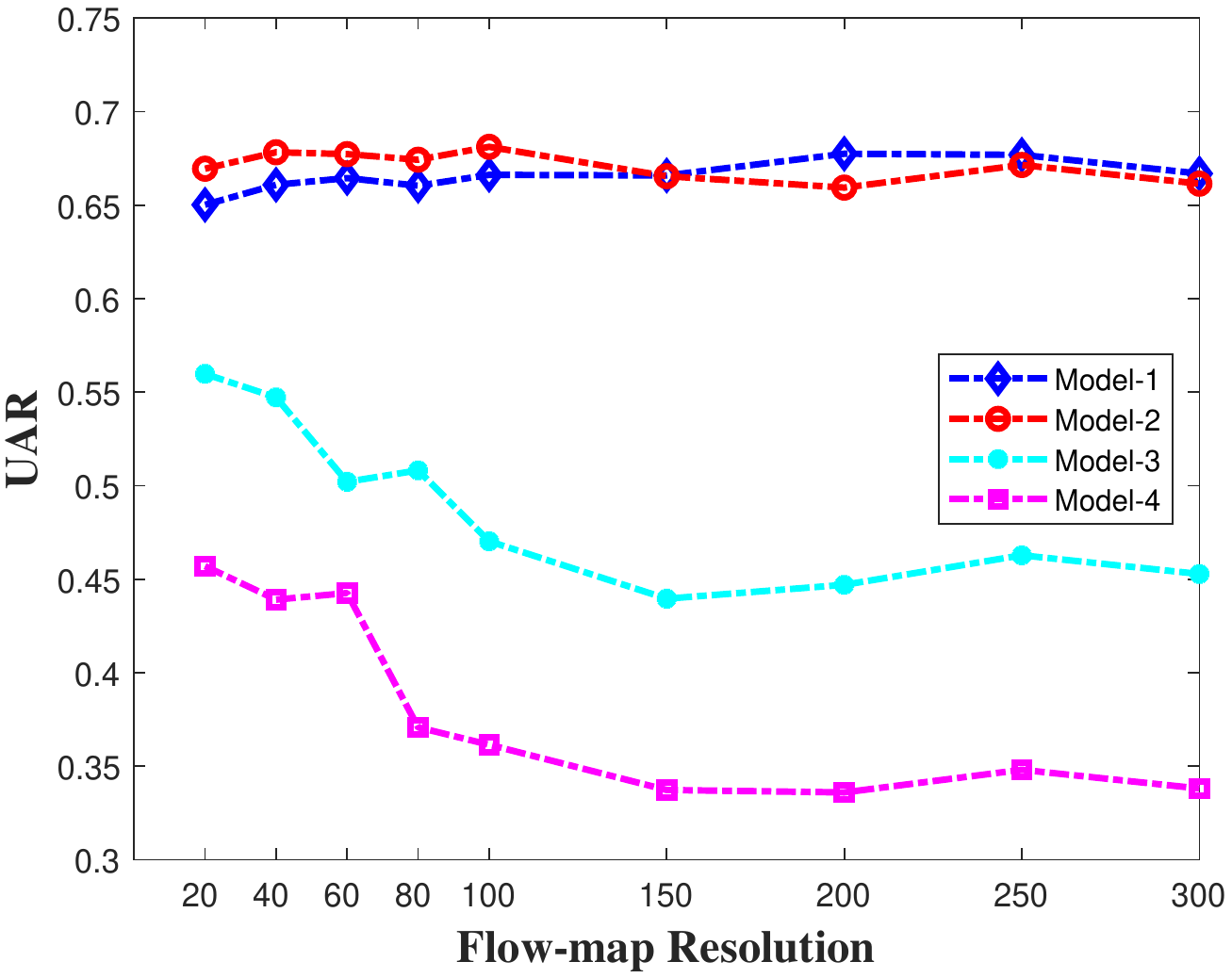}
  \caption{The UAR performances of various model architectures (Model 1: 1 conv, Model 2: 1 conv+1 rconv, Model 3: 1 conv+2 rconv, and Model 4: 1 conv+3 rconv) with different input data resolutions on composite MEGC2019 dataset.}
  \label{fig:accds}
\end{figure}
\section{Phenomenon Analysis of Learning Complexity}
\label{sec:pavm}
In this section, we will introduce the influence of learning complexity (i.e., model and input complexity) to the task of composite-database MER. Here, we employ the network architecture and input data resolution to exhibit the model and input complexity, respectively. So, we perform several deep models with different network layers and input data resolutions on the MEGC2019 dataset, implying various learning complexities for composite-database task. Then we quantitatively and qualitatively analyze their influence for this task.

Here, we employ four baseline models (Model 1, 2, 3 and 4) by using one standard convolutional layer, several recurrent convolutional layers as well as one classification layer. One recurrent layer can be unfolded into several standard convolutional layers and usually regarded as a set of several standard convolutional layers with shared parameters \cite{Xia2019Spatiotemporal}. Intuitively, the larger the model indexing is, the deeper the model is. These models have same parameter configurations but merely different architectures. For simplification, we only focus on the number of convolutional layers but ignore the classification layer in the following content. \textbf{Model 1} has only 1 convolutional layer (1 conv). \textbf{Model 2} uses 1 convolutional layer and 1 recurrent convolutional layer (1 conv+1 rconv), which has only one more layer than Model 1. \textbf{Model 3} has 1 convolutional layer and 2 recurrent convolutional layers (1 conv+2 rconv) and has one more layer than Model 2. \textbf{Model 4} uses 1 convolutional layer and 3 recurrent convolutional layers (1 conv+3 rconv) and has one more layer than Model 3. Besides, we employ several resolutions (e.g., 20, 40, 60, 80, 100, 150, 200, 250, and 300) of model input to observe their influences. Following \cite{Liong2019Shallow, Liu2019Neural, Zhou2019Dual, Xia2019Cross}, we use the optical flows as the model input and the flow-map resolution determines the input complexity, which would be introduced in Section \ref{sec:md}.

\begin{figure}[t]
  \centering
  \includegraphics[width=0.65\linewidth]{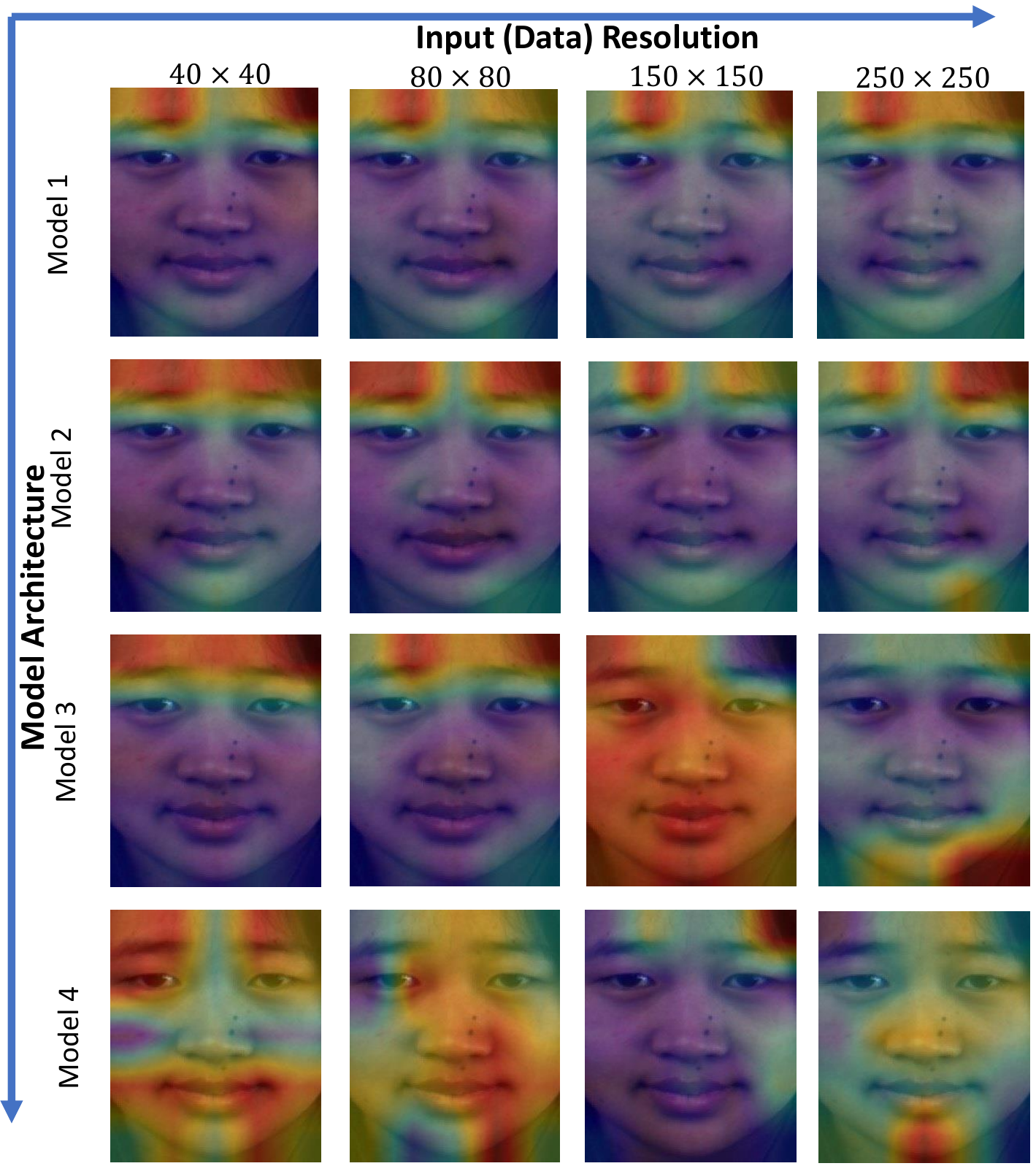}
  \caption{The class activation maps of various model architectures (Model 1: 1 conv, Model 2: 1 conv+1 rconv, Model 3: 1 conv+2 rconv, and Model 4: 1 conv+3 rconv) with different data resolutions (flow-map resolutions: $40\times40$, $80\times80$, $150\times150$ and $250\times250$) on the exemplar sample (``negative'' class, ``$sub02/EP09\_01$'' of CASME II).}
  \label{fig:visds}
\end{figure}
Fig. \ref{fig:accds} quantitatively shows the performance of various models with different-layer architectures when different-resolution flow maps are fed as the model input. The unweighted average recall (UAR) used in the challenge of MEGC2019 is also chosen as the performance evaluation. From Fig. \ref{fig:accds}, it is obviously observed that using deeper models, e.g., Model 3 and 4, have worse recognition performances compared to shallower (certain deep or shallow) models, e.g., Model 1 and Model 2. Even the architectures of Model 3 and Model 4 are much shallower than the conventional deep models, such as ResNet-101 \cite{He2016Deep} and DenseNet-169 \cite{Huang2017Densely}, they are still very easy to be distracted by the domain shift between each dataset in the task of composite-database MER. On the other side, we also observe that the model performance will be degraded with larger input resolutions (Model 2, Model 3 and Model 4). Especially, for the deeper models (Model 3 and 4), the performance begins to decrease dramatically when the resolution is bigger than $100\times100$. Oppositely, the shallower models are basically robust to the change of input resolution and can be kept almost stable for lower-resolution input. When the resolutions for shallower models are larger than a certain threshold, e.g., $40\times40$, the performance becomes better with lower-resolution input. Consequently, it indicates that suitable input resolutions (flow-map resolutions) and model architectures (network layers) are helpful to ease the degradation.

On the other side, in a qualitative view, we can exhibit the effectiveness of model learning in Fig. \ref{fig:visds}. The class activation mapping (CAM) \cite{Zhou2016Learning} is used to illustrate the focusing regions of these deep models. For the exemplar sequence, the micro-expression (``negative'' emotion) occurs in the area of subject's forehead (above the eyebrows). The shallower-architecture models successfully locate the related area for this micro-expression while the deeper-architecture models have no specially attended regions or even wrong located regions. For instance, Model 3 focuses on either the entire face (no specially attended region) or the facial periphery (wrong located region) when feeding the deep model with larger input data resolutions, i.e., $150\times150$ and $250\times250$, which is the similar situation for Model 4.

According to these two-perspective observations, it can be concluded that shallower-architecture models having less model layers on lower-resolution input (data) achieve better results for the task of composite-database MER. In other words, the model having low learning complexity can obtain better representation ability in this task. Differently, the deeper models on higher-resolution data can achieve better performance on the task of individual-database MER \cite{Gan2019Off, Xia2019Spatiotemporal}. Consequently, we can make an assumption that the model with higher learning complexity learns the inter-database noises (domain shift) more easily and becomes prone to overfitting to these between-domain noises. The learned deep models will be obstructed by mixing micro-expressions from different datasets having implicit domain shifts among them. Lower-resolution data can weaken the domain shift by a downsampling operation while the shallower-architecture model with less learnable parameters becomes less sensitive to the domain shift. Therefore, it is encouraged that shrinking model and data simultaneously are adopted on the task of composite-database MER. However, in general, both lower-resolution data and shallower-architecture model will greatly limit the representation ability of models and it will become very difficult to promote the recognition performance without adding sufficient convolutional layers and learnable parameters. So, in this paper, we propose a basic recurrent convolutional network and develop three parameter-free extension modules to further explore powerful representation ability and enable the model being robust to the domain shift.

\section{Methodology}
\label{sec:md}
\begin{figure}[t]
  \centering
  \includegraphics[width=0.95\linewidth]{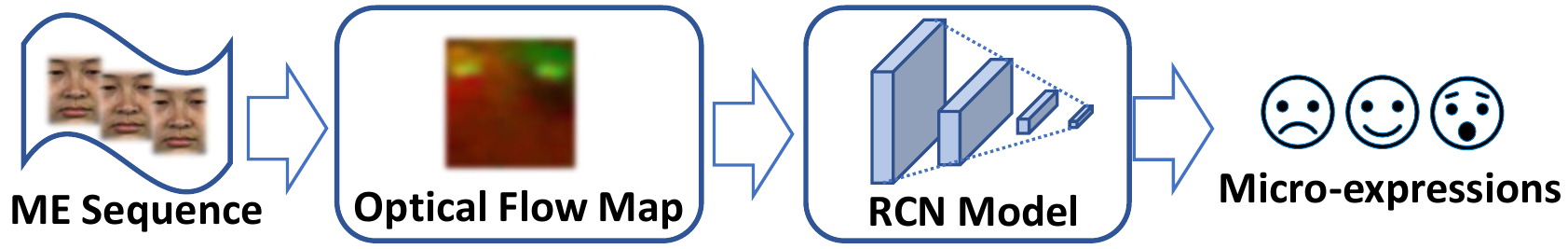}
  \caption{The diagram of our proposed method for composite-database MER.}
  \label{fig:framework}
\end{figure}

In this section, we introduce recurrent convolutional networks (RCNs) with three parameter-free modules respectively for shrinking the model and data towards the task of composite-database micro-expression recognition. Fig. \ref{fig:framework} shows the diagram of our proposed approach. In the following, the extraction of optical flow map will be introduced firstly. Then the basic architecture of RCN model will be briefly presented. Furthermore, three parameter-free modules and an automatic combination method based on NAS will further be explained in detail. The model learning for basic RCN and three modules will be finally introduced.

\subsection{Optical Flow Map Extraction}
The optical flows from the onset and apex frames are commonly used for extracting motion deformations of facial regions and can achieve good performance in subject-independent evaluation \cite{Xia2019Spatiotemporal,Liong2019Shallow, Liu2019Neural, Xia2019Cross}, which is more realistic in real-world applications. In this paper, we also employ the optical flow map as the input of our proposed model.

Before calculating the optical flow maps, the onset and apex frames need to be located for the computation. Since the onset frames have been provided in all micro-expression datasets, only apex frames need to be detected from the video sequences. Here, we use the method in \cite{Xia2019Spatiotemporal} for approximately locating the apex frames. The located onset and apex frames are represented as $I_o$ and $I_a$ respectively with the width $W$ and height $H$.

Given any location $(x,y,t)$ in a sequence, the optical flow field (containing horizontal and vertical components $V_x$ and $V_y$) of each location can be estimated by solving the following equation
\begin{equation}
I_{x}V_{x}+I_{y}V_{y}=-I_{t}
\end{equation}
where $I_x$, $I_y$ and $I_t$ are the derivatives with respect to $x$, $y$ and $t$. These gradients can be approximately calculated by using numerical partial derivatives between gray-scaled apex frame $I_a$ and onset frame $I_o$. To solve the above under-constrained problem, we employ the Lorentzian penalty optimization function \cite{Sun2010Secrets} to estimate the optical flow $V = [V_x~V_y]$, , where $V \in \mathbb{R}^{W\times H \times 2}$.

In order to measure the variation of optical flow fields, we further utilize the first-order derivatives of optical flow field $V$ (also called as optical strain) to complete the optical flow and construct a three-dimensional tensor, which is similar to an RGB image. The first-order derivative value of $V$ is calculated as
\begin{equation}
V_z=\bigg(\frac{\partial V_x}{\partial x}^2+\frac{\partial V_y}{\partial y}^2+\frac{1}{2}\big(\frac{\partial V_x}{\partial y}^2+\frac{\partial V_y}{\partial x}^2\big)\bigg)^{\frac{1}{2}}
\end{equation}
where $\frac{\partial V_x}{\partial x}$, $\frac{\partial V_y}{\partial y}$, $\frac{\partial V_x}{\partial y}$ and $\frac{\partial V_y}{\partial x}$ are the corresponding partial first-order derivatives of $V$. Combined the optical flow and its first-order variation, a three-dimensional map is generated and can be represented as $\hat{V} = [V_x~V_y~V_z]$, where $\hat{V} \in \mathbb{R}^{W\times H \times 3}$. This map will finally be used as the model input to further learn discriminative features for MER.

\subsection{Recurrent Convolutional Network}
\begin{figure}[t]
  \centering
  \includegraphics[width=0.7\linewidth]{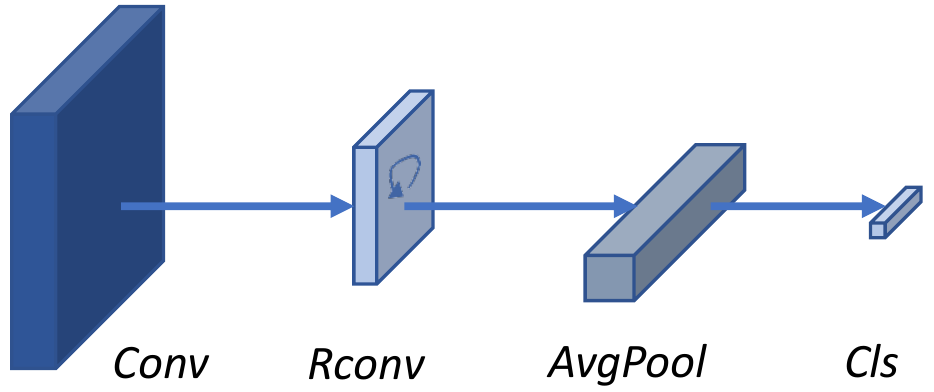}
  \caption{The basic architecture of recurrent convolutional network (RCN), which contains totally four layers.}
  \label{fig:backbone}
\end{figure}
Based on the conclusion of Section \ref{sec:pavm}, we choose to use Model 2 as the backbone of our proposed model, which pursues a trade-off between powerful representation ability and domain-shift overfitting. The architecture of Model 2 is shown in Fig. \ref{fig:backbone}, which contains one standard convolutional layer ($Conv$), one recurrent convolutional layer ($Rconv$), one average pooling ($AvgPool$) and one classification layer ($Cls$). With the input tensor $\hat{V}$, the first three layers extract visual features while the last layer recognizes micro-expressions with a probability vector $p_c\in \mathbb{R}^{C \times 1}$.

The standard convolutional layer $Conv$ performs the convolution operation for one time while the recurrent convolutional layer $Rconv$ performs the convolution operation for several times as it has a recurrent input. So the recurrent layer can be regarded as several convolutional layers by sharing parameters. To make the convolution clearer, we compare these two types of operations as follows
\begin{equation}
\begin{split}
&\mathbf{F}^c = \mathbf{W}^c * \mathbf{X} +\mathbf{b}^c \\
&\mathbf{F}_n^r = \mathbf{W}_n^r * \mathbf{X} + \mathbf{W}_{n}^r * \mathbf{F}_{n-1}^r + \mathbf{b}_n^r
\end{split}
\label{eq:reconv}
\end{equation}
where $\mathbf{X}$ denotes the final output of last layer (input of current layer). $\mathbf{F}^c$ and $\mathbf{F}_n^r$ represent the outputs (feature maps) of standard and recurrent convolution operations, respectively. $\mathbf{W}^c$ denotes the weight of standard convolution. $\mathbf{F}_{n-1}^r$ represents the output of last (time) state and $\mathbf{W}_n^r$ is the convolution weight at the current (time) state. In this context, $\mathbf{F}_0^r=\mathbf{X}$. $\mathbf{b}^c$ and $\mathbf{b}^r_n$ are the corresponding bias for standard and recurrent convolutions respectively.

Following each convolution operation, we further adopt normalization and activation operations to obtain the final feature maps of each convolutional layer. The processing procedure can be modeled in the following equation
\begin{equation}
\mathbf{X} = g(\mathbf{F})
\label{eq:normact}
\end{equation}
where $g(\cdot)$ represents the normalization and activation operations. In this paper, we employ the batch normalization and rectified linear unit (ReLU) for the processing. For decreasing the spatial resolution of feature maps, the max-pooling operations are performed on the output $\mathbf{X}$ at the end of recurrent layer.

To concatenate these features of convolutional and recurrent layers for the classification, we use the adaptive average pooling and flatten the pooling output as a feature vector, rather than using the global average pooling directly. At last, following the concatenated output, a linear classification layer is used to classify the $C$-categories micro-expressions and Softmax function is used to normalize the predicted probability among $C$-categories.

\subsection{Parameter-free Extension Modules}
\begin{figure}[t]
  \centering
  \includegraphics[width=0.7\linewidth]{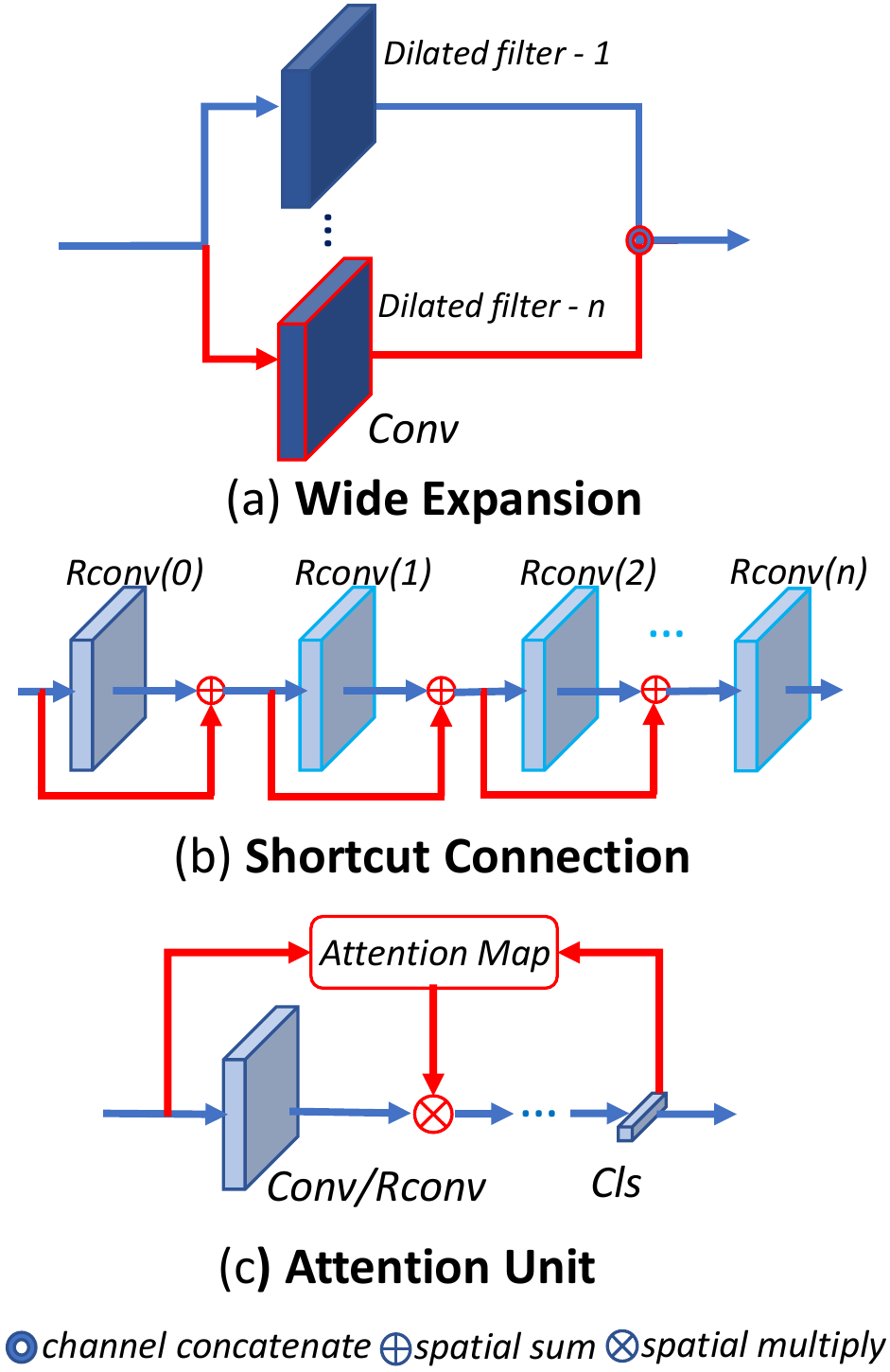}
  \caption{The schematic diagram of proposed parameter-free modules for extending basic RCN.}
  \label{fig:modules}
\end{figure}
Usually, the deep models can promote the representation ability by containing more learnable modules. However, the increase of learnable modules comes with a cost of overfitting risk for composite-database MER. So, we propose three extension modules to promote the representation ability but not introduce any more learnable parameters for avoiding model overfitting. These extension modules contain the wide-expansion, shortcut-connection and attention unit, which are successively shown in Fig. \ref{fig:modules}.

\subsubsection{Wide Expansion Extension}
The deeper models can extract multi-level features layer by layer and achieve multiple receptive fields by using pooling operations \cite{Yu2016Multi}. When the model becomes shallower and less layers are included into the end-to-end framework, less receptive fields are considered, limiting the representation ability of the model with shallower architecture. Inspired by multiple-stream deep models \cite{Khor2019Dual, Huang2019Pain}, we utilize various convolutional operations with different receptive fields in same layer to mimic different receptive fields, rather than using multiple filters with different initial weights in different layers. However, the standard convolutional filter with larger receptive field implies more learnable parameters, which is not helpful for improving the recognition performance. In this context, we employ the dilated convolution operation \cite{Yu2016Multi} to replace the standard convolution operation, which can obtain larger receptive fields with same learnable parameters of standard convolution.

The wide expansion structure, shown in Fig. \ref{fig:modules}(a), is proposed to replace the standard convolutions with dilated convolutions. In this extension, one convolutional layer in basic RCN can be expanded into different streams for performing different-size dilation operations and then combined together for outputting to proceeding layer. The first stream uses the dilated convolution with dilation size $1$ (also treated as standard convolution) and other streams utilize the dilated convolution with various dilation sizes (larger than 1). Then these outputs of multiple streams are then concatenated in channels to generate wider feature maps as the final output of one convolutional layer. The layer after wide expansion has the same channel number as the standard layer of RCN while it contains more receptive fields in one layer. The dilated convolution operations can be summarized as
\begin{equation}
\mathbf{F}(p) = \sum_{s+l\cdot k= p} \mathbf{W}(k)*\mathbf{X}(s)
\end{equation}
where $p$ represents arbitrary location of feature map, $s$ and $k$ denotes the corresponding locations in the filter and input weight matrices. When $l = 1$, this operator performs the standard convolution; otherwise, it performs the dilated convolution. In this context, we will replace the first convolution layer of RCN in a wide expansion way and denote it as \textbf{RCN-W}.

\subsubsection{Shortcut Connection Extension}
The shortcut connections, e.g., ResNet \cite{He2016Deep} and DenseNet \cite{Huang2017Densely}, have been proven to be effective to incorporate deeper architectures for preventing gradient explosion. To exploit the shortcut connection for improving the representation ability, we propose to add one more shortcut connection to the recurrent layer (second layer in RCN). By unfolding the recurrent layer into several standard convolutional layers, the shortcut connection for different-state convolutional layers can be illustrated in Fig. \ref{fig:modules}(b), which is actually an expansion of Eq. \ref{eq:reconv}. In practice, for computing efficiently, we employ an individual convolutional layer for first layer (initial state $0$) while parameter-shared convolutional layers (states $1 \sim n$) are used to represent the recurrent calculation.

With a shortcut connection (red lines in Fig. \ref{fig:modules}(b)), the input and output of previous-state network are added as the input of current-state network. The output feature map is then changed to the following
\begin{equation}
\begin{split}
&\mathbf{F}_0^r =  \mathbf{W}_{0}^r * \mathbf{X} + \mathbf{b}_0^r, \; n=0 \\
&\mathbf{F}_n^r = \mathbf{W}_n^r * \hat{\mathbf{F}}_{n-1}^r + \mathbf{W}_{n}^r * \mathbf{F}_{n-1}^r + \mathbf{b}_n^r, \; n \geq 1
\end{split}
\label{eq:dc}
\end{equation}
where $\hat{\mathbf{F}}_{n-1}^r$ represents the input of unfolded convolution operation in state ($n-1$). Similar to the backbone framework, the final output of recurrent layer with more shortcut connections is further fed into the normalization and activation functions by Eq. \ref{eq:normact}.  In this context, we will replace the recurrent layer of RCN in the shortcut connection way and denote it as \textbf{RCN-S}.

\subsubsection{Attention Unit Extension}
The previous works \cite{Wang2015Rec,Liu2016A,Xia2019Spatiotemporal} have shown that using partial facial regions achieves better recognition performance in the appearance based methods. However, the geometric based methods have not used region based strategies. Besides, in the last section, it can be observed that the learned models may focus on the micro-expression-unaware regions when using the optical flow to learn the representations. This would disturb the learning by the domain noises. So, in this paper, we propose a soft attention mechanism to strengthen the influence of micro-expression-aware regions by a parameter-free attention unit, which is shown in Fig. \ref{fig:modules}(c).

Normally, the attention unit can be constructed by one or several convolutional layers to learn the region-weighted parameters. However, in the task of composite-database MER, the learnable parameters in attention unit may make the model being prone to learning the domain shift. So it is necessary to develop the parameter-free attention unit. Whereas, the conventional saliency map methods, such as class activation map \cite{Zhou2016Learning}, can visualize the weighted regions but is a type of post hoc processing technique. Inspired by CAM \cite{Zhou2016Learning}, we develop a parameter-free attention unit which can be integrated into the end-to-end RCN framework.

Given the input $\mathbf{X}$ of convolutional or recurrent layer, the attention map $\mathbf{M}$ is estimated by
\begin{equation}
\mathbf{M} = \sum_{i=1}^{C}\sum_{j=1}^{N} \mathbf{W}_i \odot d(\mathbf{X}_j)
\end{equation}
where ``$\odot$'' denotes the element-wise multiplication between two matrices and $d(\cdot)$ represents the downsampling operation. $\mathbf{W}_i$ represents the weight matrix corresponding to $i$-th class and needs to be reshaped into a matrix from the weight vector of classification layer. $\mathbf{X}_j$ denotes the $j$-th channel of feature maps of $\mathbf{X}$. $C$ and $N$ denote the number of classes and feature maps, respectively. The estimated attention map is then multiplied with the output of convolutional or recurrent layer for assigning soft weights on spatial locations. Here, we integrate this unit with the recurrent layer individually following the connection way and denote it as \textbf{RCN-A}.

\subsubsection{Module Integration}
For three extension modules, there are many feasible ways to integrate these modules into RCN. For instance, the wide expansion and attention unit can also be integrated into any unfolded convolutional layer in recurrent layer. Besides, these three modules can be further combined together into the basic RCN simultaneously. These extensions cannot be easily determined in a fast way. To address this issue,  we turn to neural architecture search (NAS) and provide an automatic way to determine the combination way. Specifically, a gradient-based NAS strategy \cite{Liu2018Darts} is employed to search the integration position of each extension in a continuous space.

Instead of building the search space with different convolution operations, we construct it with different function layers, including Conv and Rconv layers. Given the feature $F_k$ from the input of $k$-th convolutional layer, we extract the output representation $F_{k+1}$ at $(k + 1)$-th feature map with the function modules we choose. Just as the one-shot NAS \cite{Liu2018Darts}, all the function modules are paralleled and the weighted sum of their outputs is treated as the output $F_{k+1}$. Eventually, based on the contribution of each module, we choose the best one with largest weight. Mathematically, that is
\begin{equation}
\label{eq:weighted}
\mathbf{F}_{k+1} = \sum_{i=1}^{M}\frac{\alpha_{k+1,i}}{\sum_{j}^{M} \alpha_{k+1,j}} \mathcal{M}_i(\mathbf{F}_{k})
\end{equation}

Here, we assume there are $M$ modules in our search space and $M_i$ denotes the $i$-th function module. Then $\alpha_{k+1,i}$, which works as the architecture parameter, is its corresponding parameter at the $k + 1$ state. Then the problem here is to search a set of parameters $\alpha \in \mathit{R}^{n \times M}$ for a network with $n$ convolutional layers (including the unfolded convolutional layers in RCN) so that $\alpha$ minimizes the loss on the validation data. Since the network parameters are interdependent with the architecture parameters, we alternatively update them so that the optimal architecture is finally found.

\subsection{Model Learning}
The basic recurrent convolutional network and the extension models with three modules mainly have three learnable parts, containing several convolutional layers and one classification layer. So we employ the stochastic gradient descent with momentum to train these models by a cross-entropy loss function. The cross-entropy loss $\mathcal{L} $ is calculated as
\begin{equation}
\begin{split}
\mathcal{L} = \sum_i\Big(-\alpha_i log(\mathbf{p}_t^i) \Big) \\
\mathbf{p}_t = \mathbf{p}_c^{\mathbf{y}}(1-\mathbf{p}_c)^{1-\mathbf{y}}
\end{split}
\end{equation}
where $\mathbf{y}$ is the label vector and its arbitrary element $y_i \in\{0,1\}$, and $\alpha_i$ is the weight of sample $i$ and set to $1$ in this context.

\section{Experiments}
\label{sec:ex}
In this section, we explain the implementation details, the composite datasets and protocols we used, and the results and corresponding analyses.

\subsection{Implementation Details}
\begin{table}[t]
\centering
\caption{The detailed configuration of our proposed models.}
\label{tab:rcns}
\scalebox{1.0}{
\centering
\begin{tabular}{|c|c|}
\hline
Layers	&Configurations \\ \hline \hline
Input	&Tensor: $W\times H \times 3$ \\ \hline
\multirow{2}{*}{Conv}	&$standard:~k:3$; $p:1$; $s:3$	\\
		&$dilated:~k:3$; $d:1,2,3$; $p:1,2, 3$; $s:3$	\\ \hline
\multirow{3}{*}{Rconv}	&$1~feed\text{-}forward$:  $k:1$, $p:0$, $s:1$ \\
		&$3~(shortcut)~recurrents$:  $k:3$, $p:1$, $s:1$	\\
		&$MaxPool$, $k:2$, $s:2$	\\ \hline
AvgPool&Adpative to $K \times K$ \\ \hline
Classification	&$K\times K \times M \times C$ \\ \hline
Output	&$C$ categories	\\ \hline \hline
\multicolumn{2}{|l|}{All the outputs of convolutional layers contain $M$ feature maps.}	\\
\multicolumn{2}{|l|}{$H, W$ - height and width of flow-maps, $C$ - number of classes.} \\
\multicolumn{2}{|l|}{$K$ - output size of average pooling, $k$ - filter or pooling size.} \\
\multicolumn{2}{|l|}{$d$ - dilation size, $p$ - padding size, $s$ - stride size.}	\\ \hline
\end{tabular}}
\end{table}

The detailed configurations of recurrent convolutional networks are shown in Table \ref{tab:rcns}, which contains the backbone model and three modules. Some important parameters (e.g., number of feature maps and size of average pooling) for promoting representation ability are investigated by grid searching in Section \ref{sec:pa} while other parameters (e.g., dilation size and stride size) are fixed for all models. In this paper, we employ three dilation sizes (totally three streams) and equally assign channels for each stream in wide expansion extension. Apart from the varying dilation sizes for RCN-W, all parameters, e.g.,  four states in one recurrent layer, are set consistent in all models.

\begin{figure}[t]
  \centering
  \includegraphics[width=1.0\linewidth]{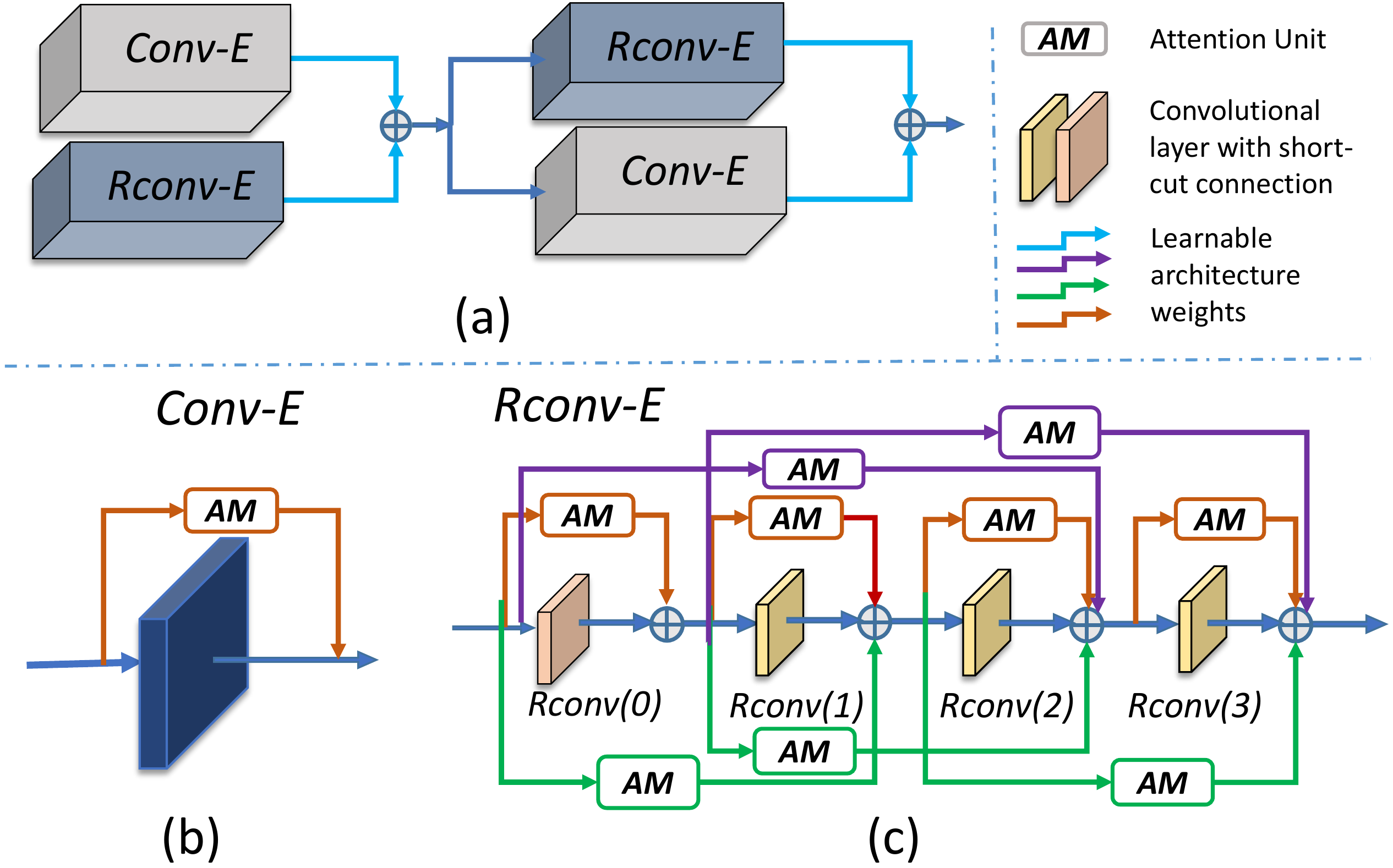}
  \caption{Illustration of our search space for model integration.}
  \label{fig:nas}
\end{figure}

For the module integration, we explore the optimal way for the combination of our models and also for the location of the attention units. The whole search space is illustrated in Fig. \ref{fig:nas}(a), where we search the best combination order by paralleling them in each computation node (hidden state). For each of these modules, as shown in Fig. \ref{fig:nas} (b) and (c), we also insert the attention unit, which is the one in Fig. \ref{fig:nas}(c) and here we simplify it for making the figure clearer. For the \textit{Rconv} network, we extend it and explore which layer is the best place for our attention unit. Here we implement one (in red), two (in green) and three (in purple)-hop attention at different layers and build a continuous search space by giving each of them an architecture weight. Then, a gradient-based NAS method is employed to find the best locations or combination. Finally, we choose the top three architectures based on their weights.

For learning parameters, the momentum is set to 0.9 and weight decay 0.0005 in stochastic gradient decent (SGD) with momentum. The learning rate is set to $10^{-4}$. The stopping criterion for SGD loss is set to $0.5$ for iterations and maximum iteration number is set to 500, respectively. For each convolutional layer, the Dropout operation with a ratio of $0.5$ is used to avoid overfitting when training. For fast implementation, we utilize the library PyTorch1.2 and train the proposed models on a GPU cluster (six Tesla K80s).

\subsection{Composite Dataset and Protocol}
In the second facial micro-expression grand challenge (MEGC2019), three commonly-used datasets, i.e., SMIC \cite{li2013spontaneous}, CASME II \cite{yan2014casme} and SAMM \cite{Davison2018SAMM}, are merged into a composite dataset, in which three overlapping labels are adopted for the recognition task. In details, ``negative'' micro-expression class groups disgust, sadness, fear, contempt and anger together as one emotion class while ``positive'' class consists of happiness emotion class. Meanwhile, ``surprise'' class is retained on its own due to the uniqueness of the emotion. The samples with the labels different in three datasets are discarded and finally $442$ samples from 68 subjects are reserved for constructing a composite dataset.

These three datasets in MEGC2019 have similar eliciting conditions and settings which let participants undergo high emotional arousal and suppress their facial expressions by watching highly emotional clips with a punishment threat and highly emotional clips. But the following characteristics are different for each dataset:
\begin{itemize}
\item SMIC comprises spontaneous micro-expressions from 164 samples of 16 subjects. Most subjects are Asians, and some are from other ethnicity. The camera recording is 100 fps high-speed and has averaged $160\times140$ resolution for facial regions.
\item CASME II remains spontaneous micro-expressions from 145 samples of 24 subjects. All subjects are Chinese. The camera for recording has 200 fps high-speed and has averaged $210\times200$ resolution for facial regions.
\item SAMM has spontaneous micro-expressions from 133 samples of 28 subjects. The subjects are very diverse and come from various ethnicity. The camera for recording has 200 fps high-speed and has averaged $230\times210$ resolution for facial regions. However, it also employs an array of LEDs to avoid flickering and only obtain gray-scale video frames.
\end{itemize}
From above descriptions, it can be easily observed that these three datasets have obvious domain shift between each other.

Since the subject-independent evaluation protocol, i.e., leave-one-subject-out (LOSO), is becoming main-stream for evaluating MER task, all experiments in this context are performed under LOSO protocol for evaluating the baseline and our proposed methods, following the settings in MEGC2019. For fairly comparison, we adopt both unweighted average recall (UAR) and unweighted F1-score (UF1) as evaluation metrics, which can measure class-balanced performance. Assume that $TP$, $FP$ and $FN$ are the true positive, false positive and false negative, respectively. The UAR (also called as unweighted accuracy) is calculated by $UAR = \frac{1}{C}\sum_{c=1}^{C}\frac{TP_c}{N_c}$ where $TP_c$ and $N_c$ are the number of true positives and all samples in $c$-th class. The $UF1$ is computed as $UF1= \frac{1}{C}\sum_{i=c}^{C}\frac{2P_c\times R_c}{P_c+R_c}$, where $P_c = \frac{TP_c}{TP_c+FP_c}$ and $R_c = \frac{TP_c}{TP_c+FN_i}$ for $c$-th class.

\subsection{Parameter Analysis and Ablation Study}
\label{sec:pa}
In this section, we investigate the impact of important parameters and components in our proposed method, including the size of average pooling, the number of feature maps, three modules and their combinations, and motion magnification. For better clarity, the basic model and models with three individual modules, i.e., wide expansion, shortcut connection and attention unit, are abbreviated as \textbf{RCN}, \textbf{RCN-W}, \textbf{RCN-S}, and \textbf{RCN-A}. The basic experimental setups use the $60\times60$ flow-map, $5\times5$ average pooling and $16$ feature maps. In the following experiments, each parameter will be changed individually in every experiment while others would use the basic values. All these methods are evaluated in the UAR metric.

\subsubsection{The Impact of Flow-map Resolution}
\begin{figure}[t]
  \centering
  \includegraphics[width=0.65\linewidth]{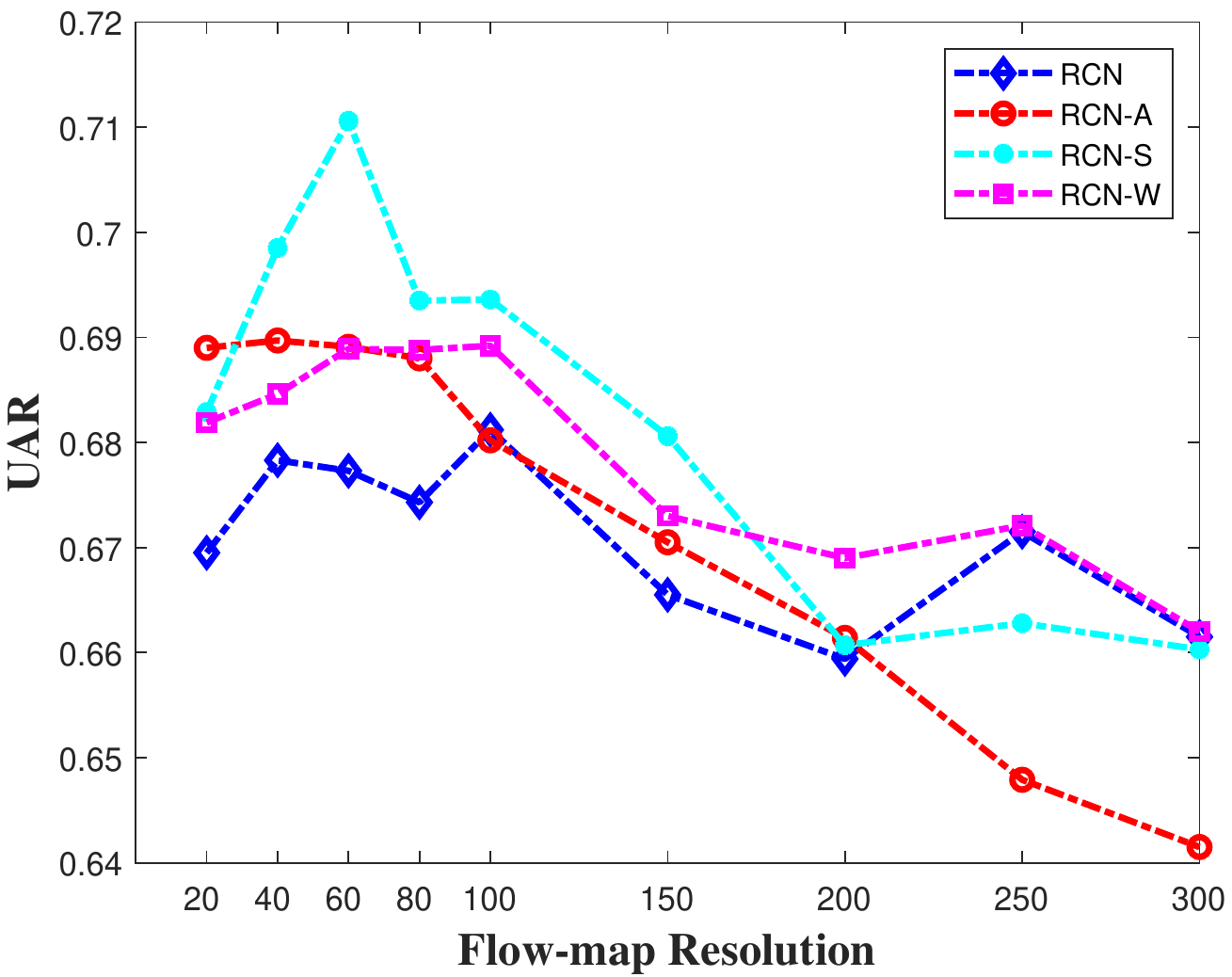}
  \caption{The UAR performance of proposed models on composite dataset MEGC2019 under different flow-map resolutions.}
  \label{fig:pmds}
\end{figure}
Fig. \ref{fig:pmds} shows the UAR performance of proposed models with various flow-map resolutions on composite dataset (MEGC2019) in the protocol of LOSO. Combined with the results from Fig. \ref{fig:accds}, we can observe that even using the proposed models, the performances are still decreased slightly by using higher flow-map resolutions. Especially, after the map resolutions are higher than $100 \times 100$, all models begin to become degraded and achieve similar representation ability even with three extension modules. When the resolution is $60 \times 60$, these three parameter-free modules obtain the best gain compared to the basic RCN model. So, in the following experiments, we adopt two flow-map resolutions, i.e., $60 \times 60$ and $100 \times 100$, as our basic flow-map resolutions for analyzing the impact of pooling sizes and number of feature maps.

\subsubsection{The Impact of Pooling Size}
\begin{figure}[t]
  \centering
  \subfloat[$60\times60$]{\includegraphics[width=0.5\linewidth]{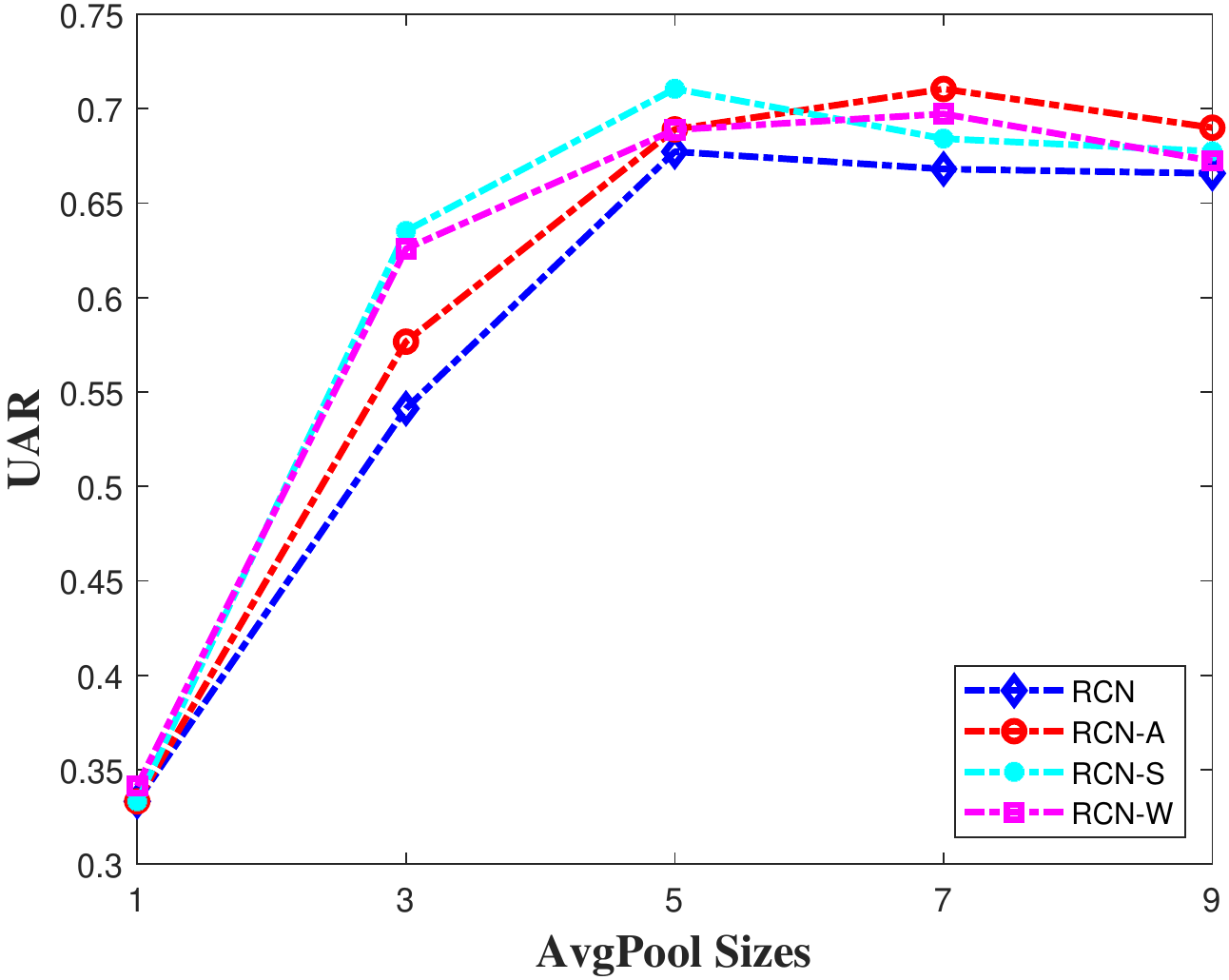}}
  \subfloat[$100\times100$]{\includegraphics[width=0.5\linewidth]{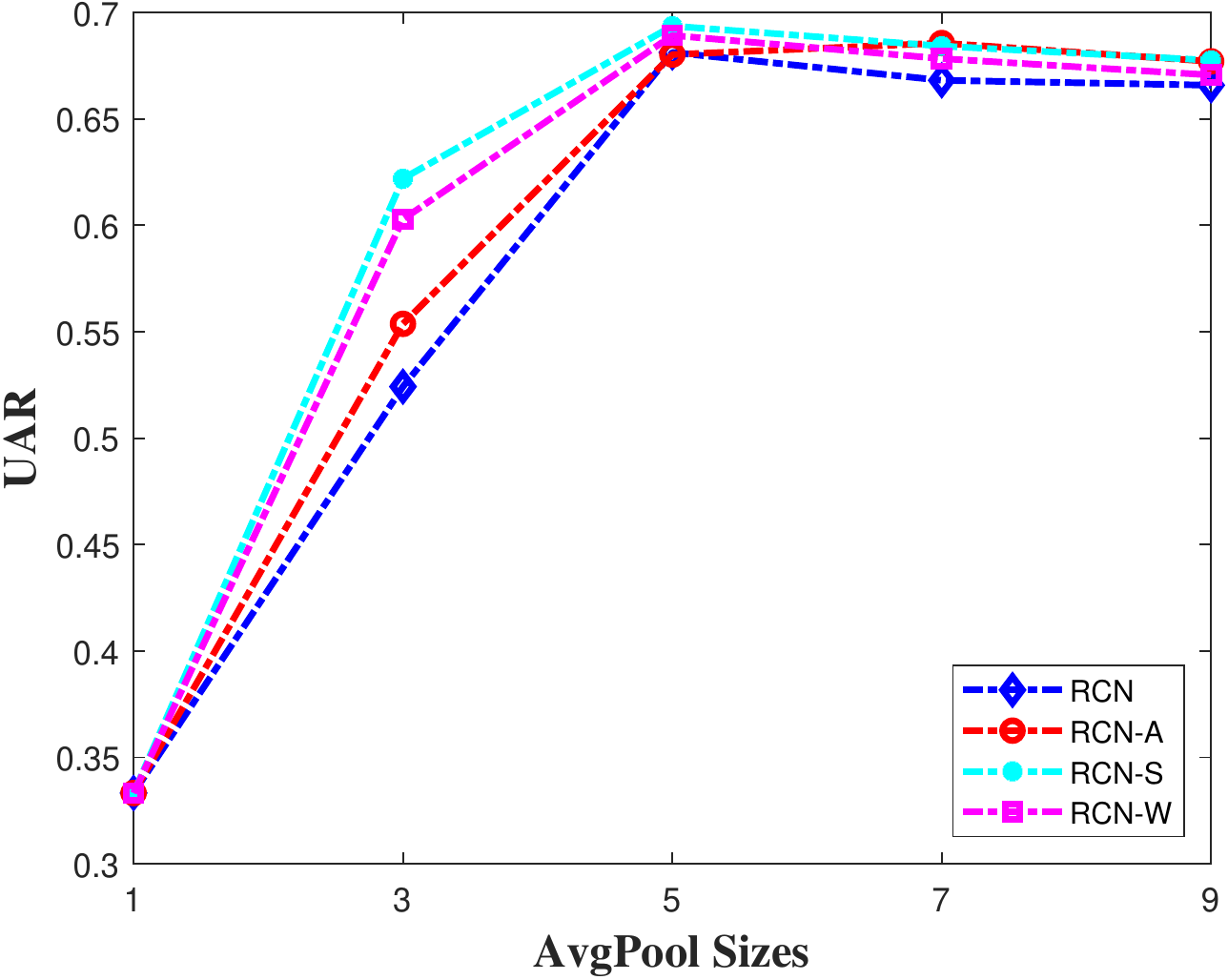}}
  \caption{The UAR performance of different average pooling sizes on MEGC2019 with flow-map resolutions of $60\times60$ and $100\times100$.}
  \label{fig:ks}
\end{figure}
Since we use the adaptive average pooling to replace the global average pooling as the last pooling operation, its pooling size will affect the representation ability of proposed models by determining the dimensionality of classification layer. In Fig. \ref{fig:ks}, the impact of using different pooling sizes is evaluated by UAR. Larger pooling sizes of adaptive average pooling imply more learnable parameters for the following classification layer (one linear classification layer). As illustrated in Fig. \ref{fig:ks}, it is easily observed that the performance with suitable pooling sizes (e.g., $5\times5$ or $7\times7$) can achieve better performance for proposed models. And we can see that their extension modules can promote the representation ability slightly compared to the basic RCN for all pooling sizes.

\subsubsection{The Impact of Feature Map}
\begin{figure}[t]
  \centering
  \subfloat[$60\times60$]{\includegraphics[width=0.5\linewidth]{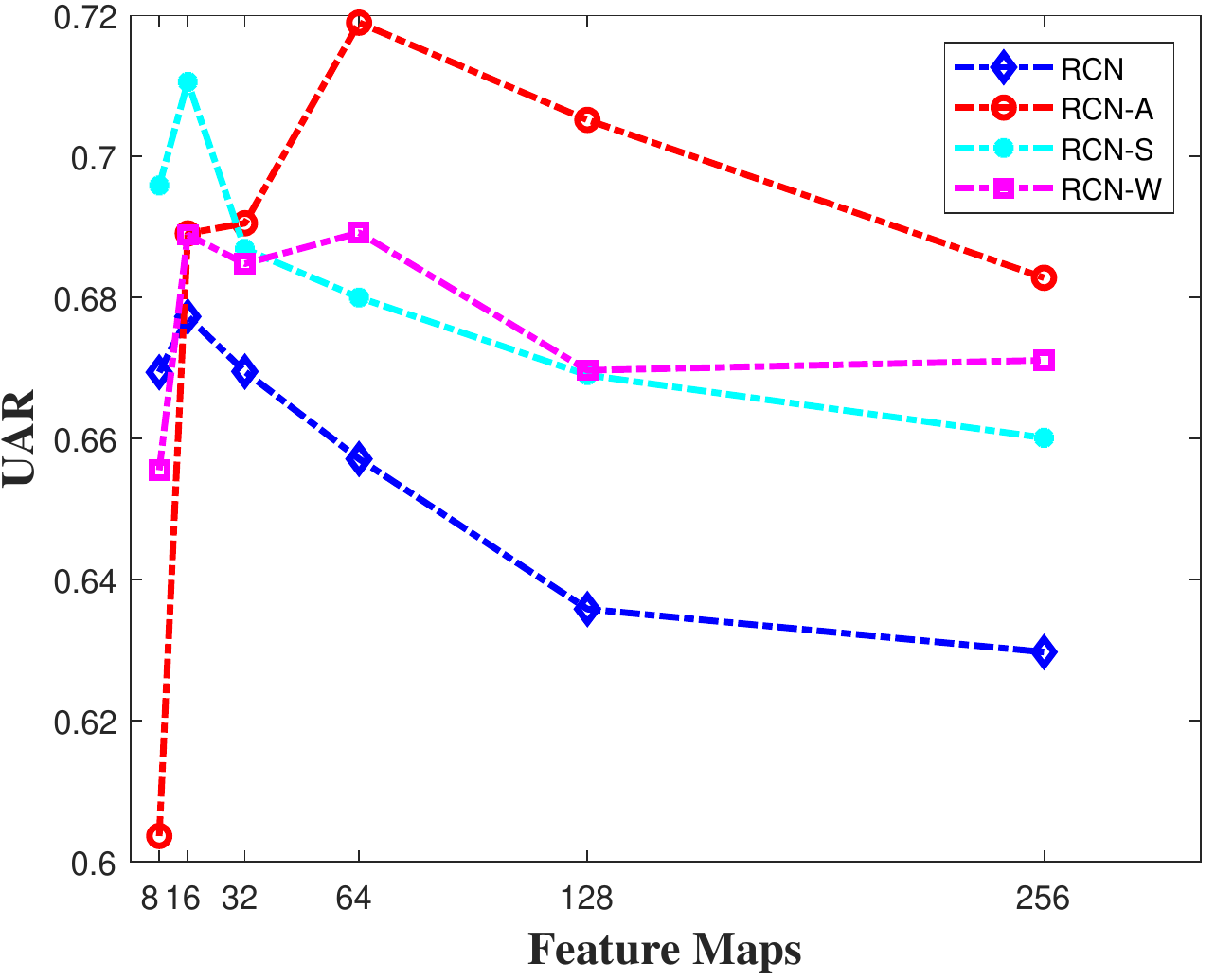}}
  \subfloat[$100\times100$]{\includegraphics[width=0.5\linewidth]{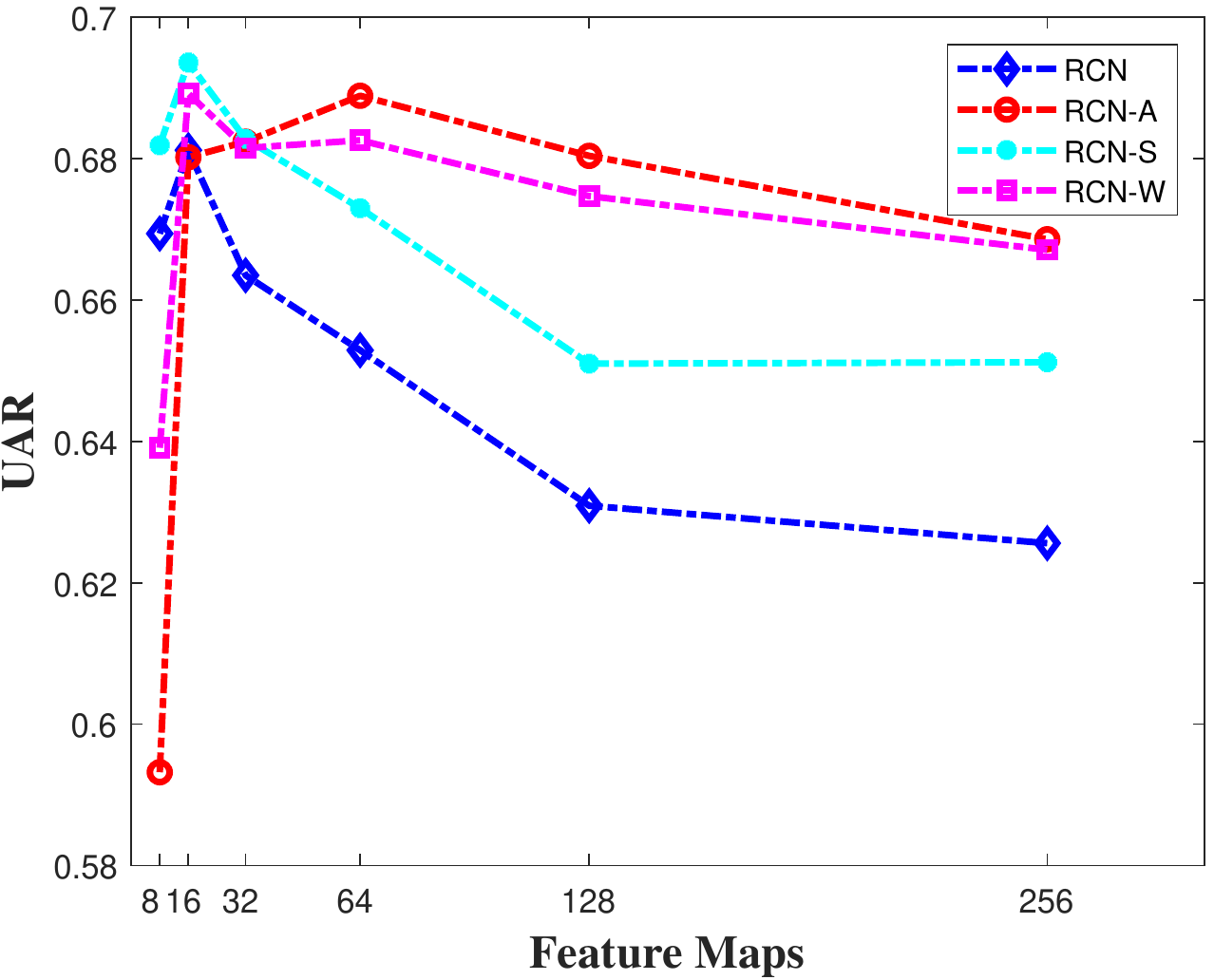}}
  \caption{The UAR performance of different numbers of feature maps on MEGC2019 with flow-map resolutions of $60\times60$ and $100\times100$.}
  \label{fig:fm}
\end{figure}
The impact of using different feature maps on various models in convolutional layers is shown in Fig. \ref{fig:fm}. More feature maps imply more learnable parameters for convolutional layers in all models. From the results of Fig. \ref{fig:fm}, it can be concluded that the models with parameter-free modules can obviously improve the performance of the basic RCN model in most cases. However, these three extension modules illustrate different trends of using various feature maps and achieve best performance by using different feature maps. RCN-S performs better with less feature maps while RCN-A and RCN-W perform better with more feature maps. Whereas, all models may degrade when much more feature maps are employed. Although much more feature maps in convolutional layers mean more powerful representation ability for describing quick subtle changes, it is easier to be disturbed by domain shift of composite dataset and affect this ability.

\subsubsection{The Impact of Combination Ways}
\begin{table}[t]
\centering
\caption{The UAR performance of alternately combining three modules under the LOSO protocol on MEGC2019 (the top two results are labeled in bold).}
\label{tab:altercom}
\scalebox{1.0}{
\centering
\begin{tabular}{|c|c|c|c|}
\hline \hline
\multicolumn{4}{|c|}{LOSO Protocol}	\\ \hline \hline
\multicolumn{4}{|c|}{Flow-map Resolution: $60\times60$, Pooling Size: $5\times5$}	\\ \hline
Approaches	&$fm=16$ &$fm=32$	&$fm=64$ 	\\ \hline
RCN	&0.6773 &0.6695 &0.6571 \\ \hline
RCN-A &0.6891 &\textbf{0.6869} &\textbf{0.7055} \\ \hline
RCN-S &\textbf{0.7106} &0.6800 &0.6801 \\ \hline
RCN-W &0.6889 &0.6848 &0.6892 \\ \hline
RCN-P	&0.6826 &0.6757 &0.6643  \\ \hline
RCN-C	&0.6829 &0.6692 &0.6578 \\ \hline
RCN-F &\textbf{0.6948} &\textbf{0.7052}&\textbf{0.6918} \\ \hline
\multicolumn{4}{|c|}{Flow-map Resolution: $60\times60$, Pooling Size: $7\times7$}	\\ \hline
Approaches	&$fm=16$	&$fm=32$	&$fm=64$ 	\\ \hline
RCN			&0.6724 &0.6638 &0.6461 \\ \hline
RCN-A &0.6892 &0.7052 &\textbf{0.7190} \\ \hline
RCN-S &\textbf{0.7106} &\textbf{0.7146} &0.6848 \\ \hline
RCN-W &0.6901 &\textbf{0.7100} &0.6815 \\ \hline
RCN-P	&0.6892 &0.6862 &0.6673  \\ \hline
RCN-C	&0.6796 &0.6787 &0.6703 \\ \hline
RCN-F &\textbf{0.6915} &0.7005&\textbf{0.6901} \\
\hline \hline
\multicolumn{4}{r}{$fm$ -  feature maps.}
\end{tabular}}
\end{table}
Each combination way is attached with a weight and searched in the MER task. For simplification, only top three searched results are discussed here. Top-1 architecture inserts the attention unit between the first convolutional layer with wide expansion and recurrent layer with the shortcut-connection in a cascade way.  Top-2 architecture inserts the attention unit with the initial state ($Rconv(0)$) of recurrent layer and then combines the first convolutional layer of wide expansion and recurrent layer with the shortcut-connection together. The searched result of top-3 architecture uses the wide expansion for first convolutional layer ($Conv$) and the attention unit paralleled with the shortcut-connection recurrent layer. We choose these top three searched combination models for investigating the impact of combining three extension modules, which are denoted as \textbf{RCN-C}, \textbf{RCN-F} and \textbf{RCN-P}, respectively.

Table \ref{tab:altercom} shows the UAR performance of these combination models and compares with basic models employed individual modules. Finally, we choose the RCN-F as the final combination model because of two reasons. Firstly, it can outperform other two combination ways (i.e., RCN-P and RCN-C) as it mostly exploits the strengths of three modules, which integrates three modules almost without any overlapping (each module operates on different layers). Secondly, it becomes more insensitive (robust) to the learnable parameters (e.g., feature maps and pooling sizes) compared to the models by using individual modules.

\subsubsection{The Impact of Motion Magnification}
\begin{table}[t]
\centering
\caption{The UAR performance of all models using motion magnification under the LOSO protocol on MEGC2019 (the top two results are labeled in bold).}
\label{tab:umm}
\scalebox{1.0}{
\centering
\begin{tabular}{|c|c|c|}
\hline \hline
\multicolumn{3}{|c|}{LOSO Protocol}	\\ \hline  \hline
\multicolumn{3}{|c|}{Flow-map Resolution: $60\times60$}	\\ \hline
Approaches	&without MM	&with MM 	\\ \hline
RCN			&0.6773 &0.6562  \\ \hline
RCN-A			&0.6891 &0.6041 \\ \hline
RCN-S			&\textbf{0.7106} &0.6854 \\ \hline
RCN-W			&0.6889 &0.6540 \\ \hline
RCN-F			&\textbf{0.6915} &0.6761 \\ \hline
\multicolumn{3}{|c|}{Flow-map Resolution:$100\times100$}	\\ \hline
Approaches	&without MM	&with MM 	\\ \hline
RCN			&0.6594 &0.6154  \\ \hline
RCN-A			&0.6614 &0.5792 \\ \hline
RCN-S			&0.6607 &0.6199 \\ \hline
RCN-W			&\textbf{0.6690} &0.6337 \\ \hline
RCN-F			&\textbf{0.6642} &0.6267 \\
\hline \hline
\multicolumn{3}{r}{MM - motion magnification.}
\end{tabular}}
\end{table}
As the motion magnification procedure has shown the performance promotion for deep models in the individual-database task \cite{Wang2018Micro,Xia2018Spontaneous,Xia2019Spatiotemporal,Liu2019Neural,Khor2019Dual}, we also investigate the influence of motion magnification in the composite-database task. By adding data processing with a motion magnification (MM) procedure \cite{Wu2012Eulerian}, the proposed models using basic feature maps and pooling sizes are compared with or without MM procedure. To fairly compare with all models, the same experimental settings are applied to all models. The comparison results are shown in Table \ref{tab:umm}. Different from the individual-database task, it can be concluded that using MM procedure would decrease the performance of all deep models in the composite-database task. The reason may be that the MM procedure will magnify the domain noises even it also magnifies the motion deformations. However, the proposed models with powerful representation ability become easier to be affected by the amplified noises.

\subsubsection{The Evaluation on Partially Composited Datasets}
\begin{table}[t]
\centering
\caption{The UAR performance of proposed methods under the LOSO protocol on partially composite datasets (the top two results are labeled in bold).}
\label{tab:pcd}
\scalebox{1.0}{
\centering
\begin{tabular}{|c|c|c|c|}
\hline \hline
\multicolumn{4}{|c|}{LOSO Protocol}	\\ \hline \hline
\multicolumn{4}{|c|}{Flow-map Resolution: $60\times60$, Feature aps: 16, Pooling Size: $5\times5$}	\\ \hline
	&SMIC-CASME II&CASME II-SAMM&SMIC-SAMM\\ \hline
RCN &0.6524 &0.6163 &0.5678 \\ \hline
RCN-A &\textbf{0.7087} &0.6309 &\textbf{0.6489} \\ \hline
RCN-S &0.6757 &\textbf{0.7022} &\textbf{0.6196} \\ \hline
RCN-W &0.6754 &\textbf{0.7102} &0.6036 \\ \hline
RCN-F &\textbf{0.6873} &0.6950 &0.6166 \\ \hline
\multicolumn{4}{|c|}{Flow-map Resolution: $60\times60$ , Feature Maps: 64, Pooling Size: $7\times7$}	\\ \hline
	&SMIC-CASME II&CASME II-SAMM&SMIC-SAMM\\ \hline
RCN	&0.6737 &0.6716 &0.5456 \\ \hline
RCN-A &\textbf{0.7202} &0.7052 &\textbf{0.6394} \\ \hline
RCN-S &0.6964 &\textbf{0.7491} &0.5835 \\ \hline
RCN-W &0.7067 &\textbf{0.7055} &0.5779 \\ \hline
RCN-F &\textbf{0.7254} &0.6954&\textbf{0.5978} \\
\hline \hline
\end{tabular}}
\end{table}
\begin{table*}[t]
\centering
\caption{The UAR and UF1 performance of different methods under the LOSO protocol on MEGC2019 (the top two results are labeled in bold).}
\label{tab:sota_loso}
\scalebox{1.0}{
\centering
\begin{tabular}{|c|c|c|c|c|c|c|c|c|}
\hline \hline
\multirow{2}{*}{Approaches}	&\multicolumn{2}{c|}{MEGC2019}&\multicolumn{2}{c|}{SMIC}	&\multicolumn{2}{c|}{CASME II}	&\multicolumn{2}{c|}{SAMM}  	\\ \cline{2-9}
                   		&UAR	&UF1	&UAR	&UF1	&UAR	&UF1	&UAR	&UF1 \\
\hline \hline
LBP-TOP	&0.5785 &0.5882 &0.5280 &0.2000 &0.7429 &0.7026	&0.4102 &0.3954	\\ \hline
Bi-WOOF \cite{Liong2019Shallow}	&- &0.6296 &- &0.5727 &- &0.7805 &- &0.5211	 \\ \hline \hline
ResNet18	&0.5629 &0.5894 &0.4327 &0.4609 &0.6136 &0.6248 &0.4359 &0.4762\\ \hline
DenseNet121	&0.3414 &0.4253 &0.3334 &0.4604 &0.3518 &0.2909 &0.3374 &0.5645\\ \hline
OFF-ApexNet \cite{Liong2019Shallow}		&-	&0.7196 &- &\textbf{0.6817} &- &\textbf{0.8764} &- &0.5409\\ \hline
STSTNet* \cite{Liong2019Shallow} &0.6724 &0.7118	&0.5995 &0.5969 &0.7962 &0.8519 &0.5813	 &0.7019\\  \hline
NMER* \cite{Liu2019Neural}	&0.5936 &0.5936 &0.5555 &0.5607 &0.6929 &0.7624 &0.4894 &0.6389\\ \hline
Dual-Inception* \cite{Zhou2019Dual}	&0.6858 &0.7188 &0.6149 &0.6104 &0.8132 &0.8263 &0.5927 &0.6520\\ \hline
CapsuleNet \cite{Van2019Capsulenet}	&0.6506 &0.6520 &0.5877 &0.5820 &0.7018 &0.7068 &0.5989 &0.6209\\ \hline
DCN-DB \cite{Xia2019Cross}	&0.6037 &0.5979 &0.6327 &0.6330 &0.5797 &0.5596 &0.5234 &0.5303\\
\hline \hline
RCN-A (Ours)		&\textbf{0.7190} &\textbf{0.7432} &0.6441 &0.6326 &\textbf{0.8123} &0.8512 &\textbf{0.6715} &\textbf{0.7601}\\ \hline
RCN-S (Ours)		&0.7106 &\textbf{0.7466} &\textbf{0.6572} &0.6519 &0.7914 &0.8360 &0.6565 &\textbf{0.7647}\\ \hline
RCN-W (Ours)		&\textbf{0.7100} &0.7422&\textbf{0.6600} &\textbf{0.6584} &\textbf{0.8131} &0.8522 &0.6164 &0.7164\\ \hline
RCN-F (Ours)		&0.7052 &0.7164 &0.5980 &0.5991 & 0.8087 &\textbf{0.8563} &\textbf{0.6771} &0.6976	\\
\hline \hline
\multicolumn{9}{r}{* -  re-implemented in PyTorch1.2.}
\end{tabular}}
\end{table*}

Except the entire composite dataset, we also evaluate the proposed models on partially composite datasets. Since the MEGC2019 contains three individual datasets, we combine each two of them to generate three partially composite datasets, i.e., SMIC-CASME II, CASME II-SAMM and SMIC-SAMM. Table \ref{tab:pcd} shows the UAR performance of the proposed models evaluated on partially composite datasets. For each partially composite dataset, different configurations for proposed methods are required to obtain best representations. Obviously, it can be observed that the domain shift exists between individual datasets, especially for SMIC and SAMM datasets. Moreover, our proposed extension modules and combination way can also promote the performance of basic RCN on partially composite datasets.

\subsection{Comparison to State of the Arts}
We report the comparison results of our proposed models (RCN-A, RCN-S, RCN-W and RCN-F) in LOSO protocol with all state-of-the-art approaches, including two conventional methods \cite{pfister2011recognising, Liong2018Less} and several deep methods \cite{Liong2019Shallow, Liu2019Neural, Zhou2019Dual, Van2019Capsulenet, Xia2019Cross} in Table \ref{tab:sota_loso}. For fair comparison, all methods use the optimal parameters (e.g., kernel sizes and feature maps). The statistical results on individual datasets comprising of the composite dataset are also shown in this table.

\subsubsection{Comparison Results to Conventional Methods}
LBP-TOP \cite{pfister2011recognising} is an appearance based feature while Bi-WOOF is a geometric based feature \cite{Liong2018Less}. Both of them employ the SVM as the classifier. The LBP-TOP was used as the baseline of MEGC2019 and its results were reported on the official website. The results of Bi-WOOF were reported under the same protocol and experimental settings in \cite{Liong2019Shallow}, which used TV-L1 optical flow \cite{Khor2019Dual} to compute the optical flow maps and did not use the UAR metric.

From Table \ref{tab:sota_loso}, we can see that our proposed models with three extension modules, which also belong to the geometric based method, achieve obviously better performance than these two handcrafted methods. Actually, most deep models outperform the handcrafted features in most settings. However, as the composite dataset has obvious domain shift among datasets, some much deeper models (ResNet18 and Densenet121) cannot learn better representations compared to the handcrafted methods. It indirectly verifies the conclusion of Section \ref{sec:pavm} that the deeper models may be overfitting to the domain noises. Besides, compared the LBP-TOP and CapsuleNet with Bi-WOOF and other deep methods, it shows that the geometric based methods including the handcrafted and learned models perform better than appearance based methods in most cases. Different from the appearance based method, geometric based methods may eliminate the intra-class information of each subject (e.g., identity variation) as only geometric information of subjects is reserved.

\subsubsection{Comparison Results to Deep Methods}
We choose the ResNet18 and DenseNet121\footnote{https://github.com/pytorch/vision/blob/master/torchvision/models} as our baseline for all deep models. The OFF-ApexNet was proposed for individual-database task while it was used for comparison in \cite{Liong2019Shallow}, which used a different calculation for the metric and did not share the code. So we report the results directly from \cite{Liong2019Shallow}. Since CapsuleNet \cite{Van2019Capsulenet} and DCN-DB \cite{Xia2019Cross} used the same protocol (LOSO), same metrics (UAR and UF1) and same implementation platform (PyTorch) on MEGC2019, the results for these two methods are directly reported as they publish. Since STSTNet \cite{Liong2019Shallow}, NMER \cite{Liu2019Neural} and Dual-Inception \cite{Zhou2019Dual} used different metric calculation or platforms, we implement these three methods in PyTorch and rerun the comparison experiments according to their released codes on GitHub. The parameters in STSTNet \cite{Liong2019Shallow}, NMER \cite{Liong2019Shallow} and Dual-Inception \cite{Zhou2019Dual} are same as their shared codes. For instance, 16 feature maps are used in STSTNet, pretrained ResNet18 are transferred in NMER and 1024-dimentional fully connected layer is employed in Dual-Inception. For our methods with three extension modules and their combination, we also adopt the optimal parameters for each model in this section, rather than using the fixed ones as in Section \ref{sec:pa}. The recurrent convolutional network with attention unit (RCN-A) employs $64$ feature maps and $7\times7$ pooling size; the recurrent convolutional network with shortcut connection (RCN-S) uses $16$ feature maps and $5\times5$ pooling size;  the recurrent convolutional network with wide expansion (RCN-W) employs $32$ feature maps and $7\times7$ pooling size; the recurrent convolutional network choosing from the searched results (RCN-F) uses $32$ feature maps and $5\times5$ pooling size. Our proposed models and three re-implemented methods use our optical flow extraction algorithm for obtaining required model input.

From the Table \ref{tab:sota_loso}, our proposed methods (i.e., RCN-A, RCN-S, RCN-W and RCN-F) outperform the deep methods (ResNet18, DenseNet121, OFF-ApexNet, STSTNet, NMER, Dual-Inception, CapsuleNet and DCN-DB) obviously in most cases under the LOSO protocol. For the full composite dataset, the RCNs with extension modules can achieve better performance than other deep methods. Whereas, the OFF-ApexNet and Dual-Inception employing two streams in an ensemble way can achieve better or similar performance in some sub-datasets of MEGC2019 while they get worse performance in many cases, which imply these methods do not have good generalization to all subjects of the composite dataset. Compared to NMER, CapsuleNet and DCN-DB having similar architecture complexity with ResNet18, our proposed methods can improve the performance and inhibit the domain noises by shrinking the learning complexity.  Besides, our proposed extension modules for RCN can further improve the representation ability compared to STSTNet, OFF-ApexNet and Dual-Inception. Overall, it indicates that the parameter-free modules for recurrent convolutional network are helpful to obtain better representations while avoiding the overfitting to domain shift through shrinking the learning complexity.

\subsection{Computational Time and Visualization}
\label{sec:vip}
\begin{table}[t]
\centering
\caption{The computational time (hours) for the proposed method on MEGC2019 dataset.}
\label{tab:ctime}
\scalebox{1.0}{
\centering
\begin{tabular}{|c|c|c|c|c|}
\hline \hline
\multicolumn{5}{|c|}{Training and Test Time}	\\ \hline \hline
\multicolumn{5}{|c|}{Feature maps: $16$, Pooling Size: $5\times5$}	\\ \hline
Approaches &$fs=60$ &$fs=100$	&$fs=200$ &$fs=300$ 	\\ \hline
RCN	&1.0h &1.8h &4.7h &11.6h \\ \hline
RCN-A 	&1.1h &1.9h &4.9h &12.1h \\ \hline
RCN-S 	&1.1h &1.8h &4.8h &12.1h\\ \hline
RCN-W	&1.1h &1.9h &4.9h &11.9h\\ \hline
RCN-F  &1.1h &1.8h &5.0h &12.2h\\ \hline
\multicolumn{5}{|c|}{Feature maps: $64$, Pooling Size: $9\times9$}	\\ \hline
Approaches &$fs=60$ &$fs=100$	&$fs=200$ &$fs=300$  	\\ \hline
RCN	&1.7h &2.1h &5.9h&13.5h \\ \hline
RCN-A 	&1.8h &2.2h &6.2h&13.8h \\ \hline
RCN-S 	&1.8h &2.2h &6.1h&13.9h \\ \hline
RCN-W	&1.8h &2.2h &6.2h&13.8h \\ \hline
RCN-F  &1.8h &2.2h&6.3h&13.9h \\ \hline
\hline \hline
\multicolumn{5}{r}{$fs$ - flow-map resolution, h - hour.}
\end{tabular}}
\end{table}

\begin{figure}[t]
  \centering
  \includegraphics[width=0.6\linewidth]{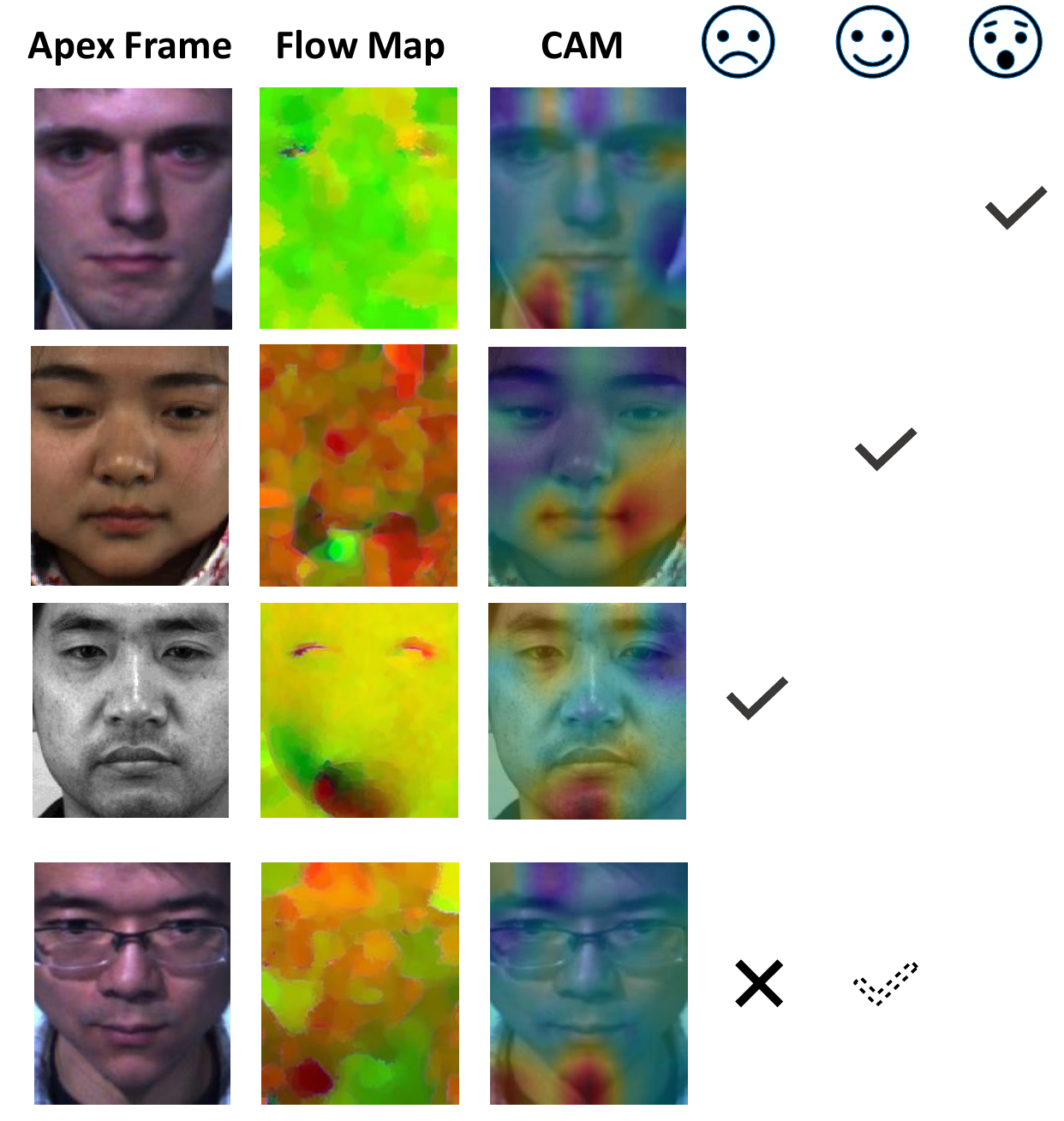}
  \caption{Exemplar recognition samples in MEGC2019 with our proposed RCN-F model for three kinds of micro-expressions (check mark denotes the corrected prediction, cross mark denotes wrong prediction and check mark with a dotted line denotes the true category).}
  \label{fig:visexam}
\end{figure}
The computational time of all proposed methods are investigated on GPU cluster (Dell PowerEdge C4130). The processor is configured as Intel Xeon E5-2650v3 at 2.30GHz with 256.0 GB RAM and the GPU accelerators are configured as NVIDIA Tesla K80 with 12.0 GB memory. In Table \ref{tab:ctime}, for each proposed model, the training and test time for all subjects ($68$ subjects) are totally included as the computational time. From the table, we can see that the higher data (flow-map) resolution will take more time to train the model, and more learnable parameters would take more computation time slightly. Hence, shrinking the data resolution will also be helpful to train the proposed models and greatly improve the learning efficiency.

Moreover, three successfully predicted examples and one example with failed prediction by our proposed RCN-F are shown in Fig. \ref{fig:visexam}. The first example (in first row) is from the subject ``s09'' of SMIC. As we can see in the figure, the surprise micro-expression usually involves the eyebrow and mouth regions, which have been captured by our proposed method. Similar things occur in the negative micro-expression and can be observed in third example (in third row), which is from the subject ``026'' of SAMM. The positive micro-expression exhibits a fast motion in the cheek region around facial mouth, which is illustrated in the second example (in second row) from the subject ``sub05'' of CASME II. These three examples demonstrate that the micro-expressions are personally diverse and make the recognition challenging. So sometimes the recognition would fail when the attention is focused on wrong regions, which is illustrated by the failed example (in fourth row) from the subject ``s19'' of SMIC.

\section{Conclusion}
\label{sec:con}
In this paper, we revealed that the learning complexity including the input data resolution and model architecture layers would affect the representation ability for composite-database micro-expression recognition. For shrinking the model and data, we further proposed recurrent convolutional networks and developed three parameter-free architectures, i.e., wide expansion, shortcut connection and attention unit. The wide expansion can promote the representation ability by obtaining more receptive fields in one convolutional layer while the shortcut connection eases the training of the recurrent layers. The attention unit is helpful to focus on the micro-expression-aware regions and extract more important information for MER. Then we presented a NAS searchable strategy to explore the potential integration ways for these three modules and chose the final architecture from the top three searched results. Extensive experiments on MEGC2019 dataset (compositing existing CASME II, SMIC and SAMM datasets) verified our analysis and showed that the proposed models with extension modules can evidently outperform the state-of-the-art approaches.

\section*{Acknowledgment}
This work is partly supported by the National Natural Science Foundation of China (Nos. 61702419, 61772419), the Natural Science Basic Research Plan in Shaanxi Province of China (No. 2018JQ6090), Tekes Fidipro program (No. 1849/31/2015) and Business Finland project (No. 3116/31/2017), Infotech Oulu, Academy of Finland ICT 2023 project (313600). As well, the authors wish to acknowledge the CSC–IT Center for Science, Finland, for computational resources.

\ifCLASSOPTIONcaptionsoff
  \newpage
\fi

\bibliographystyle{IEEEtran}
\bibliography{refs}

\begin{thebibliography}{10}
\providecommand{\url}[1]{#1}
\csname url@samestyle\endcsname
\providecommand{\newblock}{\relax}
\providecommand{\bibinfo}[2]{#2}
\providecommand{\BIBentrySTDinterwordspacing}{\spaceskip=0pt\relax}
\providecommand{\BIBentryALTinterwordstretchfactor}{4}
\providecommand{\BIBentryALTinterwordspacing}{\spaceskip=\fontdimen2\font plus
\BIBentryALTinterwordstretchfactor\fontdimen3\font minus
  \fontdimen4\font\relax}
\providecommand{\BIBforeignlanguage}[2]{{%
\expandafter\ifx\csname l@#1\endcsname\relax
\typeout{** WARNING: IEEEtran.bst: No hyphenation pattern has been}%
\typeout{** loaded for the language `#1'. Using the pattern for}%
\typeout{** the default language instead.}%
\else
\language=\csname l@#1\endcsname
\fi
#2}}
\providecommand{\BIBdecl}{\relax}
\BIBdecl

\bibitem{Takalkar2018A}
M.~Takalkar, M.~Xu, Q.~Wu, and Z.~Chaczko, ``A survey: facial micro-expression
  recognition,'' \emph{Multimedia Tools \& Applications}, vol.~77, no.~15, p.
  19301–19325, 2018.

\bibitem{li2013spontaneous}
X.~Li, T.~Pfister, X.~Huang, G.~Zhao, and M.~Pietikainen, ``A spontaneous
  micro-expression database: Inducement, collection and baseline,'' in
  \emph{IEEE International Conference and Workshops on Automatic Face and
  Gesture Recognition (FG)}.\hskip 1em plus 0.5em minus 0.4em\relax IEEE, 2013,
  pp. 1--6.

\bibitem{yan2014casme}
W.~J. Yan, X.~Li, S.~J. Wang, G.~Zhao, Y.~J. Liu, Y.-H. Chen, and X.~Fu,
  ``{CASME II}: An improved spontaneous micro-expression database and the
  baseline evaluation,'' \emph{PloS one}, vol.~9, no.~1, p. e86041, 2014.

\bibitem{Davison2018SAMM}
A.~K. Davison, C.~Lansley, N.~Costen, K.~Tan, and M.~H. Yap, ``{SAMM}: A
  spontaneous micro-facial movement dataset,'' \emph{IEEE Transactions on
  Affective Computing}, vol.~9, no.~1, pp. 116--129, 2018.

\bibitem{pfister2011recognising}
T.~Pfister, X.~Li, G.~Zhao, and M.~Pietikainen, ``Recognising spontaneous
  facial micro-expressions,'' in \emph{International Conference on Computer
  Vision (ICCV)}.\hskip 1em plus 0.5em minus 0.4em\relax IEEE, 2011, pp.
  1449--1456.

\bibitem{Ruizhernandez2013Encoding}
J.~A. Ruizhernandez and M.~Pietikainen, ``Encoding local binary patterns using
  the re-parametrization of the second order gaussian jet,'' in
  \emph{International Conference and Workshops on Automatic Face and Gesture
  Recognition (FG Workshops)}, 2013, pp. 1--6.

\bibitem{Wang2014LBP}
Y.~Wang, J.~See, R.~C.~W. Phan, and Y.~H. Oh, ``{LBP} with six intersection
  points: Reducing redundant information in lbp-top for micro-expression
  recognition,'' in \emph{Proceedings of Asian Conference on Computer Vision
  (ACCV)}, 2014, pp. 21--23.

\bibitem{Davison2014Micro}
A.~K. Davison, M.~H. Yap, N.~Costen, K.~Tan, C.~Lansley, and D.~Leightley,
  ``Micro-facial movements: An investigation on spatio-temporal descriptors,''
  in \emph{European Conference on Computer Vision (ECCV)}, 2014, pp. 111--123.

\bibitem{Huang2015Facial}
X.~Huang, S.~J. Wang, G.~Zhao, and M.~Pietikäinen, ``Facial micro-expression
  recognition using spatiotemporal local binary pattern with integral
  projection,'' in \emph{ICCV Workshop on Computer Vision for Affective
  Computing}, 2015, pp. 1--9.

\bibitem{Wang2015Rec}
S.~J. Wang, W.~J. Yan, G.~Zhao, X.~Fu, and C.~G. Zhou, ``Micro-expression
  recognition using robust principal component analysis and local
  spatiotemporal directional features,'' in \emph{European Conference on
  Computer Vision Workshops (ECCV Workshops)}, 2015, pp. 325--338.

\bibitem{Wang2015Micro}
S.~J. Wang, W.~J. Yan, X.~Li, and G.~Zhao, ``Micro-expression recognition using
  color spaces,'' \emph{IEEE Transactions on Image Processing}, vol.~24,
  no.~12, p. 6034, 2015.

\bibitem{Huang2016Spontaneous}
X.~Huang, G.~Zhao, X.~Hong, W.~Zheng, and M.~Pietikäinen, ``Spontaneous facial
  micro-expression analysis using spatiotemporal completed local quantized
  patterns,'' \emph{Neurocomputing}, vol. 175, pp. 564--578, 2016.

\bibitem{Duan2016Recognizing}
X.~Duan, Q.~Dai, X.~Wang, Y.~Wang, and Z.~Hua, ``Recognizing spontaneous
  micro-expression from eye region,'' \emph{Neurocomputing}, vol. 217, pp.
  27--36, 2016.

\bibitem{Liu2016A}
Y.~J. Liu, J.~K. Zhang, W.~J. Yan, S.~J. Wang, G.~Zhao, and X.~Fu, ``A main
  directional mean optical flow feature for spontaneous micro-expression
  recognition,'' \emph{IEEE Transactions on Affective Computing}, vol.~7,
  no.~4, pp. 299--310, 2016.

\bibitem{Li2017Towards}
X.~Li, X.~Hong, A.~Moilanen, X.~Huang, T.~Pfister, G.~Zhao, and
  M.~Pietikäinen, ``Towards reading hidden emotions: A comparative study of
  spontaneous micro-expression spotting and recognition methods,'' \emph{IEEE
  Transactions on Affective Computing}, pp. 1--1, 2017.

\bibitem{Huang2017Discriminative}
X.~Huang, S.~J. Wang, X.~Liu, G.~Zhao, X.~Feng, and M.~Pietikainen,
  ``Discriminative spatiotemporal local binary pattern with revisited integral
  projection for spontaneous facial micro-expression recognition,'' \emph{IEEE
  Transactions on Affective Computing}, vol.~PP, no.~99, pp. 1--1, 2017.

\bibitem{Liong2018Less}
S.~T. Liong, J.~See, C.~W. Phan, and K.~S. Wong, ``Less is more:
  Micro-expression recognition from video using apex frame,'' \emph{Signal
  Processing: Image Communication}, vol.~62, pp. 82--92, 2018.

\bibitem{Happy2019Fuzzy}
S.~Happy and A.~Routray, ``Fuzzy histogram of optical flow orientations for
  micro-expression recognition,'' \emph{IEEE Transactions on Affective
  Computing}, vol.~10, no.~3, pp. 394--406, 2019.

\bibitem{Zhang2019Deep}
L.~Zhang, J.~Liu, B.~Zhang, D.~Zhang, and C.~Zhu, ``Deep cascade model-based
  face recognition: When deep-layered learning meets small data,'' \emph{IEEE
  Transactions on Image Processing}, vol.~29, pp. 1016--1029, 2019.

\bibitem{Wang2018Micro}
S.-J. Wang, B.-J. Li, Y.-J. Liu, W.-J. Yan, X.~Ou, X.~Huang, F.~Xu, and X.~Fu,
  ``Micro-expression recognition with small sample size by transferring
  long-term convolutional neural network,'' \emph{Neurocomputing}, vol. 312,
  pp. 251--262, 2018.

\bibitem{Li2018Can}
Y.~Li, X.~Huang, and G.~Zhao, ``Can micro-expression be recognized based on
  single apex frame?'' in \emph{IEEE International Conference on Image
  Processing (ICIP)}.\hskip 1em plus 0.5em minus 0.4em\relax IEEE, 2018, pp.
  3094--3098.

\bibitem{Gan2019Off}
Y.~Gan, S.-T. Liong, W.-C. Yau, Y.-C. Huang, and L.-K. Tan, ``Off-apexnet on
  micro-expression recognition system,'' \emph{Signal Processing: Image
  Communication}, vol.~74, pp. 129--139, 2019.

\bibitem{Xia2018Spontaneous}
Z.~Xia, X.~Feng, X.~Hong, and G.~Zhao, ``Spontaneous facial micro-expression
  recognition via deep convolutional network,'' in \emph{International
  Conference on Image Processing Theory, Tools and Applications (IPTA)}, 2018,
  pp. 1--6.

\bibitem{Xia2019Spatiotemporal}
Z.~Xia, X.~Hong, X.~Gao, X.~Feng, and G.~Zhao, ``Spatiotemporal recurrent
  convolutional networks for recognizing spontaneous micro-expressions,''
  \emph{IEEE Transactions on Multimedia}, pp. 1--1, 2019.

\bibitem{Liong2019Shallow}
S.-T. Liong, Y.~Gan, J.~See, H.-Q. Khor, and Y.-C. Huang, ``Shallow triple
  stream three-dimensional cnn (ststnet) for micro-expression recognition,'' in
  \emph{IEEE International Conference on Automatic Face \& Gesture Recognition
  (FG)}.\hskip 1em plus 0.5em minus 0.4em\relax IEEE, 2019, pp. 1--5.

\bibitem{Liu2019Neural}
Y.~Liu, H.~Du, L.~Zheng, and T.~Gedeon, ``A neural micro-expression
  recognizer,'' in \emph{2019 14th IEEE International Conference on Automatic
  Face \& Gesture Recognition (FG)}.\hskip 1em plus 0.5em minus 0.4em\relax
  IEEE, 2019, pp. 1--4.

\bibitem{Zhou2019Dual}
L.~Zhou, Q.~Mao, and L.~Xue, ``Dual-inception network for cross-database
  micro-expression recognition,'' in \emph{IEEE International Conference on
  Automatic Face \& Gesture Recognition (FG)}.\hskip 1em plus 0.5em minus
  0.4em\relax IEEE, 2019, pp. 1--5.

\bibitem{Van2019Capsulenet}
N.~Van~Quang, J.~Chun, and T.~Tokuyama, ``Capsulenet for micro-expression
  recognition,'' in \emph{IEEE International Conference on Automatic Face \&
  Gesture Recognition (FG)}.\hskip 1em plus 0.5em minus 0.4em\relax IEEE, 2019,
  pp. 1--7.

\bibitem{Xia2019Cross}
Z.~Xia, H.~Liang, X.~Hong, and X.~Feng, ``Cross-database micro-expression
  recognition with deep convolutional networks,'' in \emph{International
  Conference on Biometric Engineering and Applications (ICBEA)}.\hskip 1em plus
  0.5em minus 0.4em\relax ACM, 2019, pp. 56--60.

\bibitem{Xia2016Spontaneous}
Z.~Xia, X.~Feng, J.~Peng, X.~Peng, and G.~Zhao, ``Spontaneous micro-expression
  spotting via geometric deformation modeling,'' \emph{Computer Vision \& Image
  Understanding}, vol. 147, pp. 87--94, 2016.

\bibitem{wang2014micro}
S.-J. Wang, W.-J. Yan, X.~Li, G.~Zhao, and X.~Fu, ``Micro-expression
  recognition using dynamic textures on tensor independent color space,'' in
  \emph{International Conference on Pattern Recognition (ICPR)}.\hskip 1em plus
  0.5em minus 0.4em\relax IEEE, 2014, pp. 4678--4683.

\bibitem{Peng2019Boost}
W.~Peng, X.~Hong, Y.~Xu, and G.~Zhao, ``A boost in revealing subtle facial
  expressions: A consolidated eulerian framework,'' in \emph{IEEE International
  Conference on Automatic Face and Gesture Recognition (FG)}, 2019.

\bibitem{Zong2018Learning}
Y.~Zong, X.~Huang, W.~Zheng, Z.~Cui, and G.~Zhao, ``Learning from hierarchical
  spatiotemporal descriptors for micro-expression recognition,'' \emph{IEEE
  Transactions on Multimedia}, vol.~PP, no.~99, pp. 1--1, 2018.

\bibitem{yao2014micro}
S.~Yao, N.~He, H.~Zhang, and O.~Yoshie, ``Micro-expression recognition by
  feature points tracking,'' in \emph{International Conference on
  Communications}.\hskip 1em plus 0.5em minus 0.4em\relax IEEE, 2014, pp. 1--4.

\bibitem{Lu2015Delaunay}
Z.~Lu, Z.~Luo, H.~Zheng, J.~Chen, and W.~Li, ``A delaunay-based temporal coding
  model for micro-expression recognition,'' in \emph{Asian Conference on
  Computer Vision Workshops (ACCV Workshops)}, 2015, pp. 698--711.

\bibitem{Xu2017Microexpression}
F.~Xu, J.~Zhang, and J.~Z. Wang, ``Microexpression identification and
  categorization using a facial dynamics map,'' \emph{IEEE Transactions on
  Affective Computing}, vol.~8, no.~2, pp. 254--267, 2017.

\bibitem{Takalkar2017Image}
M.~A. Takalkar and M.~Xu, ``Image based facial micro-expression recognition
  using deep learning on small datasets,'' in \emph{International Conference on
  Digital Image Computing: Techniques and Applications (DICTA)}, 2017, pp.
  1--7.

\bibitem{Khor2019Dual}
H.-Q. Khor, J.~See, S.-T. Liong, R.~C. Phan, and W.~Lin, ``Dual-stream shallow
  networks for facial micro-expression recognition,'' in \emph{IEEE
  International Conference on Image Processing (ICIP)}.\hskip 1em plus 0.5em
  minus 0.4em\relax IEEE, 2019, pp. 36--40.

\bibitem{Zong2018Domain}
Y.~Zong, W.~Zheng, X.~Huang, J.~Shi, Z.~Cui, and G.~Zhao, ``Domain regeneration
  for cross-database micro-expression recognition,'' \emph{IEEE Transactions on
  Image Processing}, vol.~27, no.~5, pp. 2484--2498, 2018.

\bibitem{Zong2019Toward}
Y.~Zong, W.~Zheng, Z.~Cui, G.~Zhao, and B.~Hu, ``Toward bridging
  microexpressions from different domains,'' \emph{IEEE transactions on
  cybernetics}, 2019.

\bibitem{Yap2018Facial}
M.~H. Yap, J.~See, X.~Hong, and S.-J. Wang, ``Facial micro-expressions grand
  challenge 2018 summary,'' in \emph{IEEE International Conference on Automatic
  Face \& Gesture Recognition (FG)}.\hskip 1em plus 0.5em minus 0.4em\relax
  IEEE, 2018, pp. 675--678.

\bibitem{He2016Deep}
K.~He, X.~Zhang, S.~Ren, and J.~Sun, ``Deep residual learning for image
  recognition,'' in \emph{International Conference on Computer Vision and
  Pattern Recognition (CVPR)}.\hskip 1em plus 0.5em minus 0.4em\relax IEEE,
  2016, pp. 770--778.

\bibitem{Huang2017Densely}
G.~Huang, Z.~Liu, L.~Van Der~Maaten, and K.~Q. Weinberger, ``Densely connected
  convolutional networks,'' in \emph{International Conference on Computer
  Vision and Pattern Recognition (CVPR)}, 2017, pp. 4700--4708.

\bibitem{Zhou2016Learning}
B.~Zhou, A.~Khosla, A.~Lapedriza, A.~Oliva, and A.~Torralba, ``Learning deep
  features for discriminative localization,'' in \emph{IEEE conference on
  computer vision and pattern recognition (CVPR)}, 2016, pp. 2921--2929.

\bibitem{Sun2010Secrets}
D.~Sun, S.~Roth, and M.~J. Black, ``Secrets of optical flow estimation and
  their principles,'' in \emph{International Conference on Computer Vision and
  Pattern Recognition (CVPR)}, 2010, pp. 2432--2439.

\bibitem{Yu2016Multi}
F.~Yu and V.~Koltun, ``Multi-scale context aggregation by dilated
  convolutions,'' in \emph{International Conference on Learning Representations
  (ICLR)}, 2016.

\bibitem{Huang2019Pain}
D.~Huang, Z.~Xia, L.~Li, K.~Wang, and X.~Feng, ``Pain-awareness multistream
  convolutional neural network for pain estimation,'' \emph{Journal of
  Electronic Imaging}, vol.~28, no.~4, p. 043008, 2019.

\bibitem{Liu2018Darts}
H.~Liu, K.~Simonyan, and Y.~Yang, ``Darts: Differentiable architecture
  search,'' \emph{International Conference on Learning Representations (ICLR)},
  2019.

\bibitem{Wu2012Eulerian}
H.~Y. Wu, M.~Rubinstein, E.~Shih, J.~Guttag, F.~Durand, and W.~Freeman,
  ``Eulerian video magnification for revealing subtle changes in the world,''
  \emph{ACM Transactions on Graphics}, vol.~31, no.~4, pp. 13--15, 2012.

\end{thebibliography}

\end{document}